\documentclass[journal]{IEEEtran} 

\usepackage{cite}                                
\usepackage{amsmath,amssymb,mathtools,mathrsfs}
\mathtoolsset{showonlyrefs}
\usepackage{graphicx}
\usepackage{tikz}
\usepackage{booktabs}
\usepackage{multirow}
\usepackage{tabularx}
\usepackage{array,makecell}                      
\usepackage{float}                               
\usepackage{enumitem}
\usepackage[ruled]{algorithm2e}
\usepackage[font=footnotesize]{subcaption}

\makeatletter
\@ifundefined{citet}{\newcommand{\citet}[1]{\cite{#1}}}{}
\@ifundefined{citep}{\newcommand{\citep}[1]{\cite{#1}}}{}
\makeatother

\usepackage[bookmarks=false]{hyperref}

\graphicspath{{fig/}} 

\renewcommand{\arraystretch}{1.05}   

\newcolumntype{L}[1]{>{\raggedright\arraybackslash}p{#1}}
\newcolumntype{C}[1]{>{\centering\arraybackslash}p{#1}}
\newcolumntype{R}[1]{>{\raggedleft\arraybackslash}p{#1}}
\newcolumntype{Y}{>{\raggedright\arraybackslash}X}
\newcolumntype{Z}{>{\centering\arraybackslash}X}

\allowdisplaybreaks

\usepackage{tabularx}
\usepackage{booktabs}
\usepackage{makecell}

\usepackage{amsmath,amssymb,mathtools,mathrsfs}
\mathtoolsset{showonlyrefs}

\title{A Comprehensive Survey on Surgical Digital Twin}

\author{Afsah Sharaf Khan, Falong Fan, Doohwan DH Kim, Abdurrahman Alshareef, Dong Chen,\\
Justin Kim, Ernest Carter, Bo Liu \textit{Senior Member, IEEE}, Jerzy W. Rozenblit \textit{Senior Member, IEEE},  Bernard Zeigler \textit{Fellow, IEEE}%
}

\begin{document}
\maketitle


\begin{abstract}
With the accelerating availability of multimodal surgical data and real-time computation, Surgical Digital Twins (SDTs) have emerged as virtual counterparts that mirror, predict, and inform decisions across pre-, intra-, and postoperative care. Despite promising demonstrations, SDTs face persistent challenges: fusing heterogeneous imaging, kinematics, and physiology under strict latency budgets; balancing model fidelity with computational efficiency; ensuring robustness, interpretability, and calibrated uncertainty; and achieving interoperability, privacy, and regulatory compliance in clinical environments. This survey offers a critical, structured review of SDTs. We clarify terminology and scope, propose a taxonomy by purpose, model fidelity, and data sources, and synthesize state-of-the-art achievements in deformable registration and tracking, real-time simulation and co-simulation, AR/VR guidance, edge–cloud orchestration, and AI for scene understanding and prediction. We contrast non-robotic twins with robot-in-the-loop architectures for shared control and autonomy, and identify open problems in validation and benchmarking, safety assurance and human factors, lifecycle “digital thread” integration, and scalable data governance. We conclude with a research agenda toward trustworthy, standards-aligned SDTs that deliver measurable clinical benefit. By unifying vocabulary, organizing capabilities, and highlighting gaps, this work aims to guide SDT design and deployment and catalyze translation from laboratory prototypes to routine surgical care.
\end{abstract}

\begin{IEEEkeywords}
Digital twin; surgery; robotics; simulation; real-time systems; multimodal fusion; AR/VR guidance; 
\end{IEEEkeywords}

\section{Introduction}

\subsection{Background and Significance}

\paragraph{Importance of Digital Twins in Healthcare}
Digital Twin (DT) technology represents a significant evolution in the modeling and monitoring of physical entities, systems, and processes through their high-fidelity virtual counterparts. Unlike traditional static models, DTs offer dynamic, data-driven representations that are continuously updated with real-time information. In the healthcare sector, this paradigm shift has redefined the approach to patient care by facilitating real-time modeling of physiological conditions, disease progression, and therapeutic interventions \cite{laaki2019prototyping,sun2023digital,elayan2021digital}. In parallel, perioperative care has been framed through the lens of a continuously updated “human digital twin,” emphasizing multimodal data fusion and real-time decision support across the pre-, intra-, and postoperative continuum \cite{lonsdale2022perioperative}.

By integrating diverse data sources—including genomic profiles, advanced medical imaging, and IoT sensor outputs—into comprehensive computational frameworks, DTs enable clinicians to move beyond retrospective analysis. These systems support predictive diagnostics, personalized treatment planning, and continuous health monitoring, effectively ushering in a new era of precision medicine \cite{sun2023digital,elayan2021digital,servin2024simulation}. Examples of their application are already emerging in specialized domains. In cardiology, DTs simulate electrophysiological behaviors to anticipate arrhythmia risks and fine-tune pacemaker settings, enhancing patient safety and treatment efficacy \cite{gong2023interactive,annamraju2024digital}. In oncology, they model tumor growth patterns and responses to various therapeutic regimens, thereby informing minimally invasive ablation strategies and personalized treatment pathways \cite{tai2022digital,servin2024simulation,servin2025digital}. Beyond these, vascular DTs combine mechanics-aware device/patient twins with learning-based risk indices to support endovascular planning Albertini et al. (2024), while probabilistic, risk-aware twins personalize radiotherapy fractionation in high-grade glioma \cite{chaudhuri2023predictive}.

The convergence of artificial intelligence, real-time analytics, and sophisticated simulation techniques has enabled DTs to serve as foundational tools in modern healthcare. These technologies promise not only to enhance clinical outcomes but also to reduce healthcare costs and expand access to specialized medical expertise \cite{laaki2019prototyping,sun2023digital,asciak2025digital}. The continuous evolution of DT technology underscores its central role in transforming healthcare delivery, making it an indispensable component of the emerging precision medicine landscape.

\paragraph{Emergence of Surgical Digital Twins}
Within the broad spectrum of healthcare applications, surgical practice stands out as a particularly fertile ground for the deployment of Digital Twin technology. Surgical Digital Twins (SDTs) transcend the limitations of static anatomical models by providing dynamic, patient-specific representations that encapsulate surgical procedures, operative environments, and physiological responses in real time \cite{shu2023twins,das2022toward,zhang2022artificial,asciak2025digital}. Notably, mixed-reality anatomic twins have been shown to alter heart-team strategies and unify planning in complex congenital cases \cite{lippert2024cardiac}, while “neural” reconstructions enable rapid VR-based digital twins of operating rooms for spatial planning from casual video \cite{kleinbeck2024neural}.

The foundation of SDTs lies in patient-specific anatomical modeling. These models are often constructed from preoperative imaging modalities such as CT and MRI scans and are continuously updated during procedures using intraoperative imaging techniques, optical trackers, and endoscopic systems \cite{shu2023twins,gong2023interactive,xie2024tiodt}. By maintaining an up-to-date virtual representation of the patient, SDTs enable surgeons to perform highly customized interventions with unprecedented accuracy.

In addition to anatomical modeling, SDTs facilitate comprehensive simulation of surgical workflows. These simulations encompass tool trajectories, tissue deformation, and the complex interactions between surgical instruments and biological tissues. Such capabilities enhance both preoperative planning and intraoperative guidance, enabling surgeons to anticipate procedural challenges and optimize surgical strategies \cite{shu2023twins,bjelland2022toward,hagmann2021digital,wang2025digital}. Complementary efforts in privacy-preserving workflow analysis show that de-identified “digital twin” video representations (semantic masks + depth) can enable robust event detection without exposing identities \cite{perez2025privacy}.

The integration of SDTs with hardware systems, particularly robotic surgical platforms and IoT, enabled surgical instruments—further amplifies their utility. By mirroring the actions of robotic systems, such as the da Vinci Surgical System, and leveraging sensor-enabled tools, SDTs enhance haptic feedback mechanisms and refine precision control during surgical procedures \cite{zhang2022artificial,xie2024tiodt,wang2025digital}. In training contexts, VR-based digital-twin simulators and classroom twins have demonstrated construct validity and learning gains, respectively, for robotic skills and kinematics instruction \cite{cai2023implementation,tarng2024application}.

Moreover, SDTs are instrumental in providing context-aware assistance during surgeries. Through augmented reality overlays, they highlight critical anatomical structures, offer predictive alerts for potential complications like hemorrhage or thermal injury, and even support semi-autonomous robotic maneuvers in complex surgical scenarios \cite{tai2022digital,servin2024simulation,kaliappan2024digital,servin2025digital}. Practical applications are already evident: in microwave ablation procedures for liver tumors, SDTs simulate thermal dispersion patterns to guide probe positioning and minimize collateral tissue damage \cite{tai2022digital,servin2023interactive,servin2025digital}. Similarly, in neurosurgical operations, SDTs facilitate real-time tracking of intricate maneuvers, such as skull-base drilling, thereby protecting vital neural structures \cite{shu2023twins,chen2024networking}. Across cardiovascular and endovascular domains, anatomic twins and virtual device deployment further inform strategy selection and device sizing \cite{lippert2024cardiac,albertini2024digital}.

Beyond immediate surgical applications, SDTs are reshaping the landscape of surgical education and global healthcare delivery. Immersive virtual reality environments powered by SDTs provide realistic, risk-free training platforms that circumvent the ethical and logistical constraints of traditional surgical training \cite{filippidis2024vr,ding2024towards_automation}. Furthermore, the convergence of SDTs with 5G-enabled Internet of Medical Things (IoMT) systems enhances their potential for telesurgery and remote surgical guidance, even in regions with limited bandwidth capabilities \cite{zhang2022artificial,wang2025digital,martinez2021machine}. This evolution from reactive surgical care to a proactive, predictive, and highly personalized model underscores the transformative potential of SDTs in enhancing surgical outcomes, education, and global accessibility \cite{sun2023digital,das2022toward,asciak2025digital,annamraju2024digital}. Foundational frameworks that extend industrial DT standards to neuromusculoskeletal health care also clarify domains, functions, and compliance layers for clinically deployable twins \cite{saxby2023digital}.

\subsection{Objectives of the Survey}
This survey seeks to offer a comprehensive examination of the state-of-the-art in Surgical Digital Twins, highlighting their technological underpinnings, current applications, and potential trajectories. The primary objectives of this review are twofold: to analyze the existing technological frameworks underpinning SDTs and to identify the challenges and opportunities that define their development landscape.

\paragraph{Review of Current Technologies}
The first objective is to systematically explore the technological components that constitute the foundation of SDTs. This includes an in-depth analysis of data acquisition frameworks, which are crucial for real-time model updates. These frameworks integrate a multitude of sensors, such as optical trackers, intraoperative imaging devices, physiological monitors like ECG and EMG systems \cite{elayan2021digital}, and IoT-enabled networks that facilitate seamless data collection and transmission \cite{zhang2022artificial,xie2024tiodt,wang2025digital}. The reliability and precision of these data acquisition systems are pivotal for the effective functioning of SDTs.

Equally important are the modeling and simulation techniques that underpin SDTs. This review covers a spectrum of methodologies, from biomechanical modeling approaches like finite element analysis \cite{tai2022digital,servin2024simulation}, to data-driven techniques leveraging generative adversarial networks (GANs) and reinforcement learning algorithms \cite{tai2022digital,ding2024towards_scene,servin2025digital}. The survey also examines advancements in visualization and control interfaces, particularly those employing augmented, virtual, and mixed reality technologies \cite{shu2023twins,filippidis2024vr,kaliappan2024digital}, which enhance user interaction and decision-making during surgical procedures. In parallel, we consider anatomic mixed-reality review platforms for cardiac teams \cite{lippert2024cardiac}, neural scene reconstructions for VR OR planning \cite{kleinbeck2024neural}, and privacy-preserving digital-twin video streams for workflow analysis \cite{perez2025privacy}.

Integration frameworks form the final pillar of this technological review. The role of 5G-enabled IoMT architectures \cite{tai2022digital}, large language model (LLM)-based surgical planning agents \cite{ding2024towards_scene}, and hybrid cloud-edge computing infrastructures \cite{corralacero2020digital,zhang2022artificial} are discussed in the context of their capacity to support the real-time, high-fidelity operation of SDTs. These integration mechanisms are critical for ensuring the seamless interplay between data acquisition, processing, and application within surgical settings. Complementary education-focused twins and simulators indicate pathways for workforce training and curriculum integration \cite{cai2023implementation,tarng2024application}.

\paragraph{Identification of Challenges and Opportunities}
The second objective of this survey is to identify the multifaceted challenges that hinder the widespread adoption of SDTs, as well as to highlight the emerging opportunities that could drive innovation in this field.

On the technical front, key challenges include the need for improved data interoperability and management of heterogeneous datasets \cite{das2022toward,qin2022realizing}, the demand for high-performance real-time computational systems \cite{gong2023interactive,wang2025digital}, and the ongoing quest to enhance model fidelity, especially concerning the simulation of complex soft-tissue dynamics \cite{annamraju2024digital,ding2024towards_automation}. These technical hurdles must be addressed to unlock the full potential of SDTs in clinical practice.

From a clinical and regulatory perspective, the validation and verification of SDT models present significant barriers. The stringent requirements for model accuracy, combined with complex regulatory pathways and certification processes \cite{qin2022realizing,asciak2025digital}, complicate the integration of SDTs into routine surgical practice. Ethical considerations, including algorithmic bias, patient data privacy, and informed consent \cite{ding2024towards_scene,asciak2025digital,annamraju2024digital}, further underscore the need for careful governance in the development and deployment of SDTs.

Nevertheless, the field is rich with opportunities for innovation. The synergistic integration of AI, robotics, and advanced control systems promises to enhance the intelligence and adaptability of SDTs \cite{das2022toward,xie2024tiodt,wang2025digital}. Federated learning approaches offer a promising avenue for leveraging distributed data sources while safeguarding patient privacy \cite{ding2024towards_scene}. Furthermore, the advancement of semi-autonomous and adaptive surgical systems, empowered by SDTs \cite{ding2024towards_scene,hagmann2021digital,wang2025digital}, holds the potential to revolutionize both surgical practice and education.

Through this comprehensive survey—spanning foundational theories \cite{laaki2019prototyping,sun2023digital}, applications in robotic and minimally invasive surgery \cite{xie2024tiodt}, real-time decision support systems \cite{hagmann2021digital}, and the frontier of telesurgery \cite{wang2025digital}—we aim to provide a detailed roadmap for translating the conceptual promise of Surgical Digital Twins into tangible clinical advancements.

\paragraph{Positioning vs. Existing Surveys}
Recent surveys chart the broader healthcare DT landscape and emerging directions for surgical twins, ethics, and surgical data science—Asciak et al.\ (conceptual framing of digital-twin-assisted surgery) \cite{asciak2025digital}, Katsoulakis et al.\ (health DT scoping review) \cite{katsoulakis2024digital}, Abd Elaziz et al.\ (applications/technologies/simulations in healthcare DTs) \cite{abdelaziz2024digital}, Sun et al.\ (updates and challenges) \cite{sun2023digital}, Ding et al.\ (scene understanding within surgical DTs) \cite{ding2024unifying}, and Bruynseels et al.\ (ethical implications) \cite{bruynseels2018digital}. Our survey complements these by narrowing to procedure-proximal, closed-loop SDTs intended to influence intra-operative decisions and by making explicit the modeling-to-execution-to-systems chain under realistic latency and reliability constraints.

Section \ref{section4:Mathematical Formulations} formalizes the quantitative machinery that enables an SDT to act during procedures—biophysical therapy models for ablation forecasting \cite{tai2022digital, servin2023interactive, servin2024simulation, servin2025digital}, information-gain and fusion principles for sensing \cite{blasch2021ml}, and time-critical sequential decision formulations—so that estimation, prediction, and control targets are explicit in surgical contexts.
Section \ref{section5:Connections with DEVS} connects those formulations to execution by mapping SDT components into DEVS for event-timed coordination, discussing hybrid co-simulation with continuous-time physics, and contrasting DEVS with ABM/SD/Petri-net alternatives for surgical workflows \cite{zeigler2000theory,kuruppu2025framework}.
Section \ref{section6:Robotic Surgical System and Digital Twin Integration} examines how twins couple to surgical robots and tools—including shared control, virtual fixtures, state estimation, and autonomy foundations—highlighting implications for real-time safety and human–robot interaction \cite{hagmann2021digital,wang2025digital,avgousti2020robotic,qian2023deep}.
Section \ref{section7:Communication and Simulation Tools} surfaces system constraints that bound OR deployment—edge/cloud partitioning, scheduling, bandwidth and jitter tolerance—and links the digital thread to VR/AR visualization pipelines for guidance and training, drawing on networking architectures for human DTs and latency management at the edge \cite{zhang2022artificial, filippidis2024vr, chen2024networking, bulej2021managing,yang2020bigdata}.

\begin{table*}[htbp]
\centering
\caption{Positioning relative to existing surveys}
\label{tab:positioning-surveys}
\renewcommand{\arraystretch}{1.12}
\setlength{\tabcolsep}{4pt}
\begin{tabularx}{\textwidth}{@{}p{0.12\textwidth} p{0.09\textwidth} p{0.15\textwidth} p{0.18\textwidth} p{0.4\textwidth}}
\toprule
\textbf{Category} & \textbf{Survey} & \textbf{Scope} & \textbf{Coverage highlights} & \textbf{How this survey differs} \\
\midrule
General \& surgical DT surveys &
Asciak et al.\ (2025)~\cite{asciak2025digital} &
Digital twin--assisted surgery &
Concept, opportunities, and challenges for SDTs in clinical contexts &
Develops quantitative formulations for closed-loop intra-operative use; links to DEVS and robotics/control; analyzes latency/reliability across comms/compute. \\
\addlinespace
General \& surgical DT surveys &
Katsoulakis et al.\ (2024)~\cite{katsoulakis2024digital} &
Health digital twins (scoping review) &
Landscape of health DT concepts, applications, and gaps &
Narrows to surgery; emphasizes time-critical decision loops; connects models $\rightarrow$ execution $\rightarrow$ systems constraints. \\
\addlinespace
General \& surgical DT surveys &
Abd Elaziz et al.\ (2024)~\cite{abdelaziz2024digital} &
DTs in healthcare: applications, technologies, simulations &
Enabling technologies and simulation trends across healthcare &
Frames SDTs as end-to-end systems; integrates compute/network/VR--AR with validation \& HCI for OR deployment. \\
\addlinespace
General \& surgical DT surveys &
Sun et al.\ (2023)~\cite{sun2022digital} &
Digital twin in healthcare &
Recent updates and challenges across healthcare DTs &
Procedure-proximal view with explicit performance targets (latency, observability) and robotics integration. \\
\addlinespace
General \& surgical DT surveys &
Ding et al.\ (2024)~\cite{ding2024unifying} &
DTs for surgical data science &
Geometric scene understanding within SDT representations &
Extends from perception to planning/control, risk-aware decisions, and systems (edge/cloud, VR--AR). \\
\addlinespace
General \& surgical DT surveys &
Bruynseels et al.\ (2018)~\cite{bruynseels2018digital} &
Ethics of digital twins in health care &
Privacy, autonomy, responsibility, governance &
Integrates ethics with VVUQ/validation and HCI specific to real-time surgical settings. \\
\addlinespace
Specialty-specific reviews &
Seth et al.\ (2024)~\cite{seth2024digital} &
Plastic \& reconstructive surgery &
Specialty-focused use-cases and tools &
Synthesizes cross-modal SDTs; emphasizes intra-operative decision support and emergency use. \\
\addlinespace
Specialty-specific reviews &
Corral-Acero et al.\ (2020)~\cite{corralacero2020digital} &
Precision cardiology DT &
End-to-end vision and exemplars in cardiology &
Translates end-to-end principles to surgical twins; details real-time compute/network requirements. \\
\addlinespace
Specialty-specific reviews &
Bjelland et al.\ (2022)~\cite{bjelland2022toward} &
Arthroscopic knee surgery &
Systematic review toward a knee-surgery DT &
Positions arthroscopy within a unified pipeline (data $\rightarrow$ model/sim $\rightarrow$ control/robotics $\rightarrow$ compute/network $\rightarrow$ validation/HCI). \\
\bottomrule
\end{tabularx}
\end{table*}

\begin{figure}
    \centering
    \includegraphics[width=\linewidth]{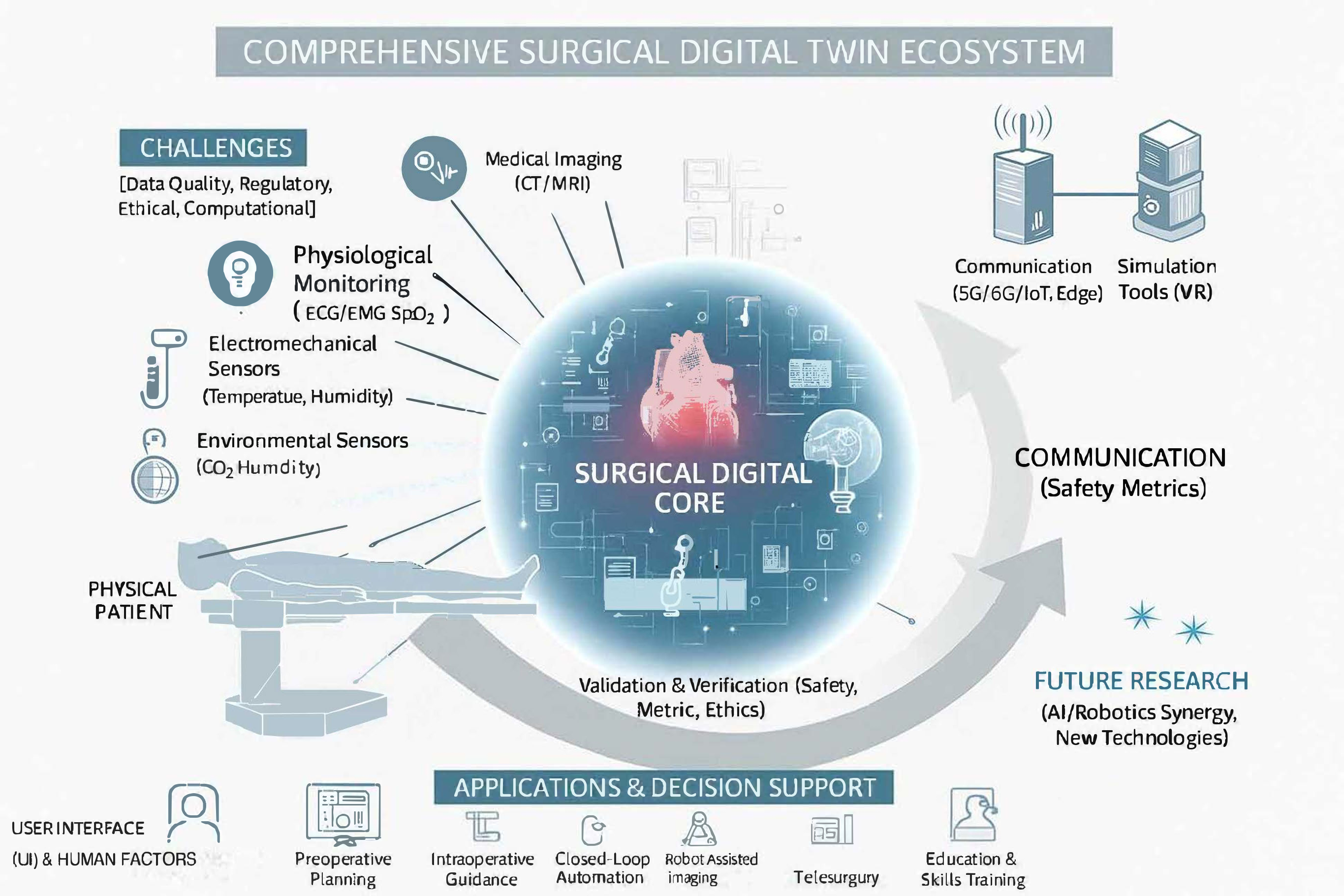}
    \caption{Digital Twin Ecosystem.}
    \label{fig:Digital Twin Ecosystem}
\end{figure}


\section{Fundamentals of Surgical Digital Twins}

\subsection{Concept and Definition of Digital Twins}

\paragraph{Origins and Evolution}
The conceptual roots of Digital Twins (DTs) can be traced to the aerospace sector, most notably NASA’s Apollo program, where ground-based simulators were utilized to mirror spacecraft behavior and support critical decision-making under variable mission conditions \cite{das2022toward}. These early mirrored systems laid the foundation for today’s real-time, virtual-physical symbiosis. The formal DT paradigm was introduced in the early 2000s, largely attributed to Michael Grieves, who characterized the DT as a persistent, digital counterpart of a physical asset, continuously synchronized via bidirectional data exchange \cite{sun2023digital}. Initially, DTs found rapid uptake in manufacturing, industrial automation, and aerospace engineering, enabling real-time monitoring, predictive maintenance, and process optimization \cite{sun2023digital,elayan2021digital,mihai2022digital}. More recently, the DT paradigm has been extended to the domains of healthcare and surgery, where patient variability, data heterogeneity, and the demand for real-time guidance create fertile ground for DT innovation \cite{elayan2021digital,das2022toward,chen2024networking}. Within healthcare, perioperative care has been explicitly framed as a “human digital twin” that continuously fuses multimodal data across pre-, intra-, and postoperative phases \cite{lonsdale2022perioperative}, and specialty frameworks have adapted industrial standards to clinical settings for neuromusculoskeletal care \cite{saxby2023digital}.

\paragraph{Core Components and Architecture}
\begin{figure}
    \centering
    \includegraphics[width=0.9\linewidth]{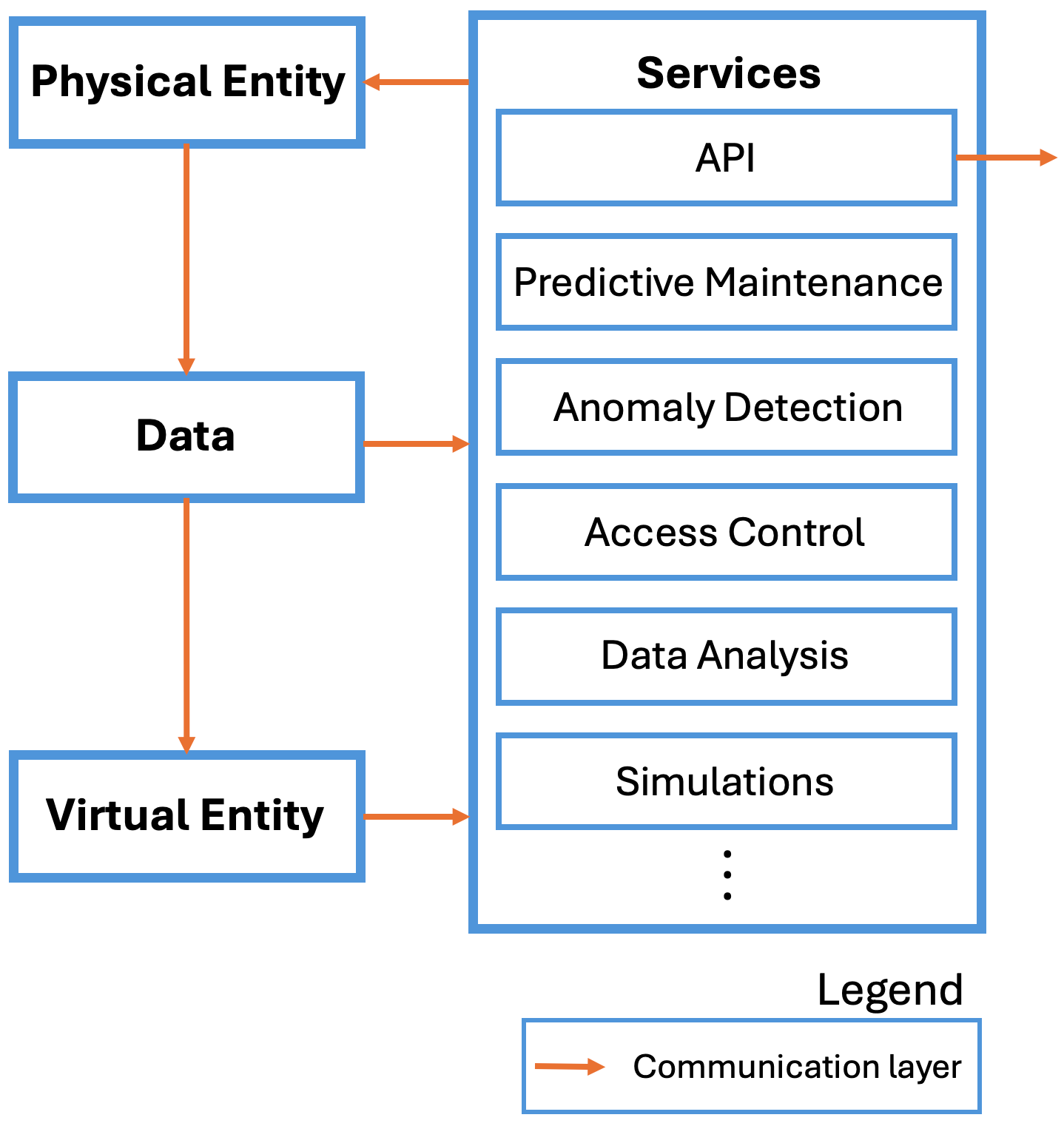}
    \caption{Five-dimensional DT architecture. \cite{mihai2022digital}.}
    \label{fig:Five-dimensional DT architecture}
\end{figure}

\begin{table*}[htbp]
\centering
\caption{Taxonomy of Surgical Digital Twins (by purpose, fidelity, data source).}
\label{tab:sdt-taxonomy}
\renewcommand{\arraystretch}{1.15}
\setlength{\tabcolsep}{4pt}
\begin{tabularx}{\textwidth}{@{} l l X X p{0.15\textwidth} @{}}
\toprule
\textbf{Axis} & \textbf{Category} & \textbf{Definition / scope} & \textbf{Typical methods / artifacts} & \textbf{References} \\
\midrule
\multicolumn{5}{@{}l}{\textit{Purpose}}\\
& Preoperative planning \& rehearsal
& Patient-specific simulation before incision; optimize strategy and anticipate complications
& CT/MRI segmentation, registration, finite-element or bioheat simulation, trajectory planning
& \citep{qin2022realizing,gong2023interactive,servin2023interactive,servin2024simulation,servin2025digital} \\
& Intraoperative guidance \& navigation
& Real-time model updates for situational awareness and targeting
& Image/pose tracking, deformable registration, AR overlays, scene tracking
& \citep{ahmed2020potential,zhang2022artificial,gong2023interactive,shi2022synergistic} \\
& Closed-loop control \& robotics/HRI
& Twin informs/receives robot commands in the loop
& Control policies, shared control, virtual fixtures, safety monitors
& Context-aware \citep{hagmann2021digital}; autonomy foundations \citep{avgousti2020robotic,qian2023deep} \\
& Telesurgery \& remote assistance
& Remote execution/supervision with synchronized twin
& QoS-aware streaming, edge/cloud split, fail-safe modes
& \citep{laaki2019prototyping,wang2025digital,suraci2025migrate,chen2024networking} \\
& Education \& skills training
& Risk-free rehearsal and assessment with realistic feedback
& VR/AR simulators, haptics, performance analytics
& \citep{filippidis2024vr,hagmann2021digital,ding2024towards_automation} \\
& Post-operative monitoring \& prognostics
& Twin tracks recovery, forecasts outcomes
& Time-series modeling, Bayesian updating, EHR integration
& \citep{sun2023digital,qin2022realizing,chen2024networking} \\
& Workflow \& OR operations
& Optimize schedules, resources, and flows around the twin
& Optimization + ML for throughput and delays
& \citep{fairley2019improving} \\
\addlinespace
\multicolumn{5}{@{}l}{\textit{Fidelity}}\\
& Geometric/anatomical
& High-resolution structure; limited physics
& Segmentation, meshing, rigid/elastic registration
& \citep{shu2023twins,qin2022realizing,gong2023interactive} \\
& Biomechanical/biophysical
& Tissue/organ mechanics and therapy physics
& FEA, bioheat/thermal dispersion, perfusion models
& \citep{tai2022digital,servin2023interactive,servin2024simulation,servin2025digital} \\
& Physiological/organ-level
& Functional physiology and electrophysiology
& Reduced-order or multi-scale organ models
& \citep{corralacero2020digital,gong2023interactive} \\
& Scene/semantic (SDS)
& Task-level understanding of surgical scene
& Geometric scene understanding, foundation models
& \citep{ding2024unifying,ding2024towards_automation} \\
& Process/Discrete-event (DEVS)
& Event-timed workflow and system dynamics
& DEVS formalism, hybrid co-simulation
& \citep{zeigler2000theory,kuruppu2025framework} \\
& Hybrid / multi-fidelity
& Mix geometric, physics, process layers as needed
& Co-simulation, surrogate/ROM, adaptive fidelity
& \citep{mihai2022digital,lu2020digital,fuller2020digital} \\
\addlinespace
\multicolumn{5}{@{}l}{\textit{Data source}}\\
& Pre-operative imaging
& Baseline anatomy and planning inputs
& CT, MRI, CTA/MRA
& \citep{qin2022realizing,gong2023interactive,servin2023interactive} \\
& Intra-operative imaging/vision
& Live updates of anatomy and tools
& Endoscopy, ultrasound, fluoro; optical tracking
& \citep{shu2023twins,ahmed2020potential,gong2023interactive,shi2022synergistic} \\
& Robotics \& kinematics
& Device/tool states for control and HRI
& Joint/pose telemetry, force/torque, video
& \citep{hagmann2021digital,wang2025digital,avgousti2020robotic} \\
& Physiological monitoring
& Signals reflecting patient state
& ECG/EMG, SpO$_2$, BP, temperature
& \citep{elayan2021digital,qin2022realizing} \\
& IoMT / wearables / environment
& Edge sensors and OR context
& Wearables, room sensors, network QoS
& \citep{tai2022digital,zhang2022artificial,wang2025digital,suraci2025migrate,chen2024networking} \\
& EHR / omics / prior data
& Longitudinal context for personalization
& EHR, labs, genomics/omics
& \citep{sun2023digital,mihai2022digital} \\
\bottomrule
\end{tabularx}
\end{table*}

A standard Digital Twin system comprises three fundamental elements: the physical entity, a high-fidelity virtual model, and the continuous data exchange that binds the two \cite{das2022toward,mihai2022digital}. The physical entity may be a medical device, organ, or patient. Its virtual model is a computationally rich, structural and functional representation, often built from multi-modal imaging, sensor data, and historical records \cite{elayan2021digital,zhang2022artificial}. The bi-directional data exchange ensures that the digital counterpart reflects the present state of the physical entity, allowing for adaptive feedback and closed-loop control \cite{sun2023digital,das2022toward}.

In surgical applications, this core architecture is augmented by advanced technologies. IoT-based sensing, including wearable and implantable devices, streams physiological signals in real time \cite{elayan2021digital,tai2022digital,zhang2022artificial}. Artificial intelligence and machine learning frameworks process this data for anomaly detection, risk prediction, and personalized recommendations \cite{sun2023digital,zhang2022artificial,asciak2025digital,mihai2022digital}. Real-time simulation engines and extended reality (AR/VR) platforms facilitate immersive surgical planning and intraoperative guidance \cite{ahmed2020potential,zhang2022artificial,asciak2025digital,gong2023interactive,shi2022synergistic}. Cloud and edge computing architectures ensure scalable, low-latency computation—critical for time-sensitive surgical workflows \cite{elayan2021digital,tai2022digital,das2022toward,chen2024networking}. Collectively, these elements converge in cyber-physical systems that enable data-driven, predictive, and adaptive surgical decision-making \cite{elayan2021digital,das2022toward,zhang2022artificial,asciak2025digital,chen2024networking}.

\paragraph{General Digital Twins}
The principles underpinning SDTs are rooted in a rich history of development from other high-stakes industries. The concept was first articulated in aerospace, where the DT served as a high-fidelity virtual counterpart for mission assurance and vehicle health management, establishing the canonical physical–virtual mirroring and telemetry loop \cite{Glaessgen2012DigitalTwin}. Early formalizations framed the DT as a persistently synchronized system designed to mitigate unpredictable emergent behavior in complex assets, with foundational guidance on virtual factory replication and lifecycle integration \cite{Grieves2017DTParadigm, Grieves2014Whitepaper}. Subsequent surveys distinguished digital models, digital shadows, and true twins by coupling strength and bidirectionality, systematizing definitions and characteristics across domains to avoid conceptual drift \cite{Kritzinger2018Survey, Barricelli2019SurveyDT, Jones2020CharacterisingDT}. At its core, simulation is positioned as the executable core of the twin—semantically rich, behaviorally faithful, and continuously state-aligned—supporting analysis, prediction, and control from design through operation \cite{Boschert2016DT, Schleich2017ShapingDT}.

As the paradigm matured, particularly in manufacturing and production, its architectural patterns became more defined. Production-oriented paradigms describe "DT shop-floor" closed loops for sensing, diagnosis, optimization, and control, which directly parallel the intraoperative feedback loops required in surgery \cite{Tao2018Shopfloor}. Reference application frameworks further decomposed data, model, and service layers, along with governance, while manufacturing CPS reference models mapped roles, interfaces, and services for scalable deployment \cite{Zheng2019Framework,Leng2021DTCPS}. Standardization efforts, notably ISO 23247, specify identifiers, information models, and interoperability patterns to enable traceable, multi-vendor DT ecosystems \cite{ISO23247-1-2021}. Furthermore, systems-engineering viewpoints embedded DTs within Model-Based Systems Engineering (MBSE) to unify requirements, verification/validation, and operational analytics for safety-critical certification—a crucial consideration for medical device development \cite{Madni2019MBSE}. Complementary CPS and production-systems reviews ground DTs in tightly coupled sensing/actuation with cyber-physical orchestration \cite{Uhlemann2017DTIndustry40,Negri2017CPSRoles}.

The technical backbone required to realize these concepts has also been extensively explored. IoT/5G surveys enumerate device-edge-cloud stacks, identity/graph management, and stream processing, which are critical to preserving twin fidelity under bandwidth and jitter constraints, complemented by multimedia/IoT perspectives on identity, synchronization, and experience integration \cite{Minerva2020DTCIoT,ElSaddik2018DTMultimedia}. Concurrently, modeling surveys emphasize surrogate models, reduced-order modeling, and uncertainty quantification as prerequisites for real-time, credible DT inference \cite{Rasheed2020Values}. Evolving typologies and implementation reviews offer practical classifications and roadmaps that translate to clinical adoption challenges \cite{Kritzinger2018Survey,Khan2020DTDefinitions}. Cross-domain exemplars beyond manufacturing, such as supply-chain observability and prescriptive analytics in agri-food networks, illustrate principles directly analogous to clinical digital threads \cite{Verdouw2021AgriFoodDT}. Ultimately, comprehensive texts consolidate these strands into a simulation-first, standards-compliant, edge–cloud architecture blueprint that underpins trustworthy DTs in safety-critical environments \cite{Tao2019BookDT}. These established principles, from architectural blueprints to technical enablers, provide a robust foundation upon which the specialized field of Surgical Digital Twins is built.

\subsection{Digital Twins in Surgery}

\paragraph{Application Domains}
DTs have quickly established relevance across diverse surgical domains. Preoperative planning is a flagship application: by constructing patient-specific digital models from imaging and sensor data, surgeons can simulate entire procedures, optimize incision sites, and anticipate complications before the first cut is made \cite{qin2022realizing,gong2023interactive,servin2023interactive,servin2024simulation}. Such simulations are further enhanced by integrating biomechanical properties and predictive analytics, as seen in the modeling of tumor ablation zones or vascular interventions \cite{gong2023interactive,servin2023interactive,servin2024simulation,servin2025digital}. Mixed-reality anatomic twins have been shown to change heart-team strategies and unify planning in congenital cardiology \cite{lippert2024cardiac}, while mechanics-aware vascular/device twins with learning-based risk indices support endovascular planning \cite{albertini2024digital}. OR-scale digital twins derived from casual video aid spatial planning and ergonomics \cite{kleinbeck2024neural}.

During operations, DTs serve as real-time intraoperative assistants, updating virtual anatomical models to reflect surgical progress and physiological shifts \cite{ahmed2020potential,zhang2022artificial,gong2023interactive}. These updated models improve navigation in complex, dynamic environments and support advanced robotic procedures \cite{ahmed2020potential,hagmann2021digital}. Postoperatively, DTs support recovery monitoring and outcome prediction, aiding in both personalized aftercare and population-level analytics \cite{sun2023digital,qin2022realizing,chen2024networking}. Privacy-preserving DT representations (semantic masks and monocular depth) further enable workflow analysis and cross-site learning without sharing raw OR videos \cite{perez2025privacy}.

Surgical education and skill development have also benefited. VR-based DT platforms provide trainees with realistic, interactive environments for rehearsal and assessment—often incorporating haptic feedback and detailed performance metrics \cite{asciak2025digital,filippidis2024vr,hagmann2021digital}. Recent systems demonstrate construct validity for RMIS skills using a VR digital-twin simulator \cite{cai2023implementation}, deploy classroom twins that synchronously couple a VR arm to a physical robot and improve learning outcomes \cite{tarng2024application}, and pilot AI “digital-twin” educator avatars to scale didactics and assessment \cite{khan202310526}. In telemedicine and remote surgery, DTs enable clinicians to perform or supervise procedures from a distance, maintaining situational awareness and control through high-fidelity digital representations and real-time data feeds \cite{laaki2019prototyping,das2022toward,wang2025digital,suraci2025migrate}. This capability has proven critical in resource-limited and emergency contexts, where expert presence is often geographically constrained \cite{laaki2019prototyping,das2022toward,suraci2025migrate}.

\paragraph{Types of Surgical Digital Twins}
Surgical DTs can be categorized along multiple axes. Patient-specific twins synthesize individual anatomical, physiological, and pathological data from high-resolution imaging and biosensors, enabling personalized simulation and intervention \cite{elayan2021digital,qin2022realizing,gong2023interactive,servin2023interactive,servin2024simulation,servin2025digital}. Procedure-centric twins focus on the surgical workflow itself, enabling the optimization and rehearsal of technical steps, device positioning, and intraoperative navigation \cite{qin2022realizing,hagmann2021digital}. Device/system twins digitally replicate surgical instruments, robotic arms, and operating rooms, supporting monitoring, maintenance, and operational analytics \cite{das2022toward,xie2024tiodt,hagmann2021digital,kleinbeck2024neural}. Hybrid twins combine patient, workflow, and device-level data, offering an integrated platform for comprehensive surgical decision support \cite{sun2023digital,das2022toward,zhang2022artificial,asciak2025digital}. Cardiovascular and endovascular exemplars highlight anatomic MR twins for team decision-making \cite{lippert2024cardiac} and device–artery interaction modeling for predictive planning \cite{albertini2024digital}.

\paragraph{Non-Robotic SDTs}
Not all surgical digital twins are coupled to a robot or mechatronic actuator. A large and rapidly maturing class of non-robotic SDTs focuses on planning, guidance, and prognosis in settings where the twin augments human decision-making without issuing low-level device commands. These systems instantiate the core DT loop—acquire, model, predict, and advise—by combining patient-specific geometry and physiology with fast simulators and visualization to influence case selection, approach planning, targeting, and risk management. In pre-procedural planning, non-robotic SDTs use imaging-derived patient models to explore device sizing, access routes, and therapy parameters before incision. Cardiovascular exemplars illustrate how patient-specific anatomical and physiological modeling can generate actionable predictions for clinical decision support and precision planning \cite{corralacero2020digital,sun2023digital,gong2023interactive,lippert2024cardiac,albertini2024digital}. For energy-based therapies, non-robotic SDTs forecast treatment effect using biophysical simulators: microwave-ablation twins predict thermal dispersion and ablation volumes to guide probe placement and energy delivery, with increasing model fidelity and data-assisted calibration reported across recent studies \cite{servin2023interactive,servin2024simulation,servin2025digital}.
During procedures, guidance-oriented SDTs update virtual anatomy and instrument context in real time to improve situation awareness and targeting—often via deformable registration and AR overlays rather than direct robot control \cite{ahmed2020potential,zhang2022artificial,gong2023interactive,shi2022synergistic}. In parallel, education and rehearsal systems leverage DTs to deliver realistic VR simulation and performance analytics for skills development and pre-case walkthroughs, improving preparedness without requiring a robotic platform \cite{filippidis2024vr,hagmann2021digital,ding2024towards_automation,cai2023implementation,tarng2024application,khan202310526}. Post-operatively, monitoring/prognostic SDTs integrate longitudinal signals (imaging, physiology, EHR) to track recovery and forecast complications, supporting personalized aftercare pathways \cite{sun2023digital,qin2022realizing,chen2024networking}.
Architecturally, non-robotic SDTs rely on the same multimodal data pipelines (pre-op CT/MRI, intra-op vision/ultrasound, physiological streams) and computational substrate (edge/cloud co-processing) as their robotic counterparts, but they prioritize latency-aware visualization, reliability, and interpretability over tight control loops. Networking and deployment choices—bandwidth, jitter tolerance, and task partitioning between edge and cloud—remain central to safe OR use and to scalable tele-guidance scenarios \cite{laaki2019prototyping,tai2022digital,das2022toward,wang2025digital,suraci2025migrate,chen2024networking}. Privacy-preserving DT encodings can further reduce data-sharing risk for workflow analytics and model training \cite{perez2025privacy}. As such, non-robotic SDTs complement robot-in-the-loop twins: they cover planning, guidance, and follow-up across specialties (e.g., cardiovascular, skull base, arthroscopy), while sharing validation, interoperability, and governance challenges identified across the DT literature \cite{corralacero2020digital,sun2023digital,shu2023twins,bjelland2022toward,qin2022realizing,mihai2022digital}.

\subsection{Overview of Existing Surgical Digital Twins}

\paragraph{Existing Surgical Digital Twins}
Research on surgical DTs has accelerated over the past decade. Hagmann et al.\ \cite{hagmann2021digital} developed a context-aware DT platform for robotic surgery training, integrating semantic data extraction and virtual fixtures to provide real-time haptic guidance. Ahmed and Devoto \cite{ahmed2020potential} underscored the value of DTs in rehearsal and intraoperative support but highlighted ongoing computational barriers to real-time tissue modeling. Qin and Wu \cite{qin2022realizing} advocated embedding DTs within computer-assisted surgery ecosystems, emphasizing the synergy between DTs, AI-driven analytics, and robotic control for enhanced precision and reduced surgeon cognitive load.
Multiple reviews stress the convergence of DTs with artificial intelligence, augmented/mixed reality, and advanced robotics as a foundation for personalized, adaptive, and precise surgical care \cite{sun2023digital,elayan2021digital, zhang2022artificial, asciak2025digital, gong2023interactive, mihai2022digital}. These integrations not only expand the reach of DTs in elective surgery but also begin to transform high-stakes, emergency procedures \cite{laaki2019prototyping, das2022toward, asciak2025digital, wang2025digital, suraci2025migrate}. Representative examples include cardiac anatomic MR twins that alter team strategy \cite{lippert2024cardiac}, vascular/device twins for predictive endovascular planning \cite{albertini2024digital}, and neural OR reconstructions for VR planning \cite{kleinbeck2024neural}.

\paragraph{Commercial and Research-based Solutions}
Translational progress is evident in both academic prototypes and commercial systems. HeartFlow Fractional Flow Reserve Computed Tomography (FFRCT) leverages patient-specific digital twin (DT) models for non-invasive assessment of coronary disease, delivering actionable insights for clinical decision-making \cite{sun2023digital}. Sim\&Cure and Cydar Medical have created platforms that operationalize digital twin (DT) principles for surgical planning and intraoperative guidance, particularly in vascular and endovascular domains \cite{qin2022realizing,gong2023interactive}. The German Aerospace Center (DLR) MiroSurge system integrates digital twin (DT) frameworks for robotic surgery training with real-time haptic feedback, illustrating the pedagogical and technical value of the digital twin (DT) approach \cite{hagmann2021digital}. FEops HEARTguide exemplifies a sophisticated digital twin (DT) modeling solution for preprocedural planning in structural heart interventions \cite{sun2023digital}. Alongside these, VR DT simulators and educator avatars aim to scale training access and objective assessment \cite{cai2023implementation,khan202310526,tarng2024application}.
Despite these successes, widespread clinical adoption remains tempered by persistent issues: model validation, interoperability, integration with legacy systems, and regulatory hurdles \cite{ahmed2020potential,asciak2025digital,gong2023interactive,martinez2021machine}. These barriers have motivated ongoing research and industry collaboration focused on establishing standards and robust, transparent evaluation frameworks \cite{das2022toward,mihai2022digital,saxby2023digital,lonsdale2022perioperative}.

\subsection{Data Management Fundamentals}
Data management lies at the core of effective surgical DT deployment. The data journey begins with acquisition, drawing on high-resolution imaging (CT, MRI), streaming physiological data from wearable and intraoperative sensors, and rich patient records \cite{sun2023digital,elayan2021digital,tai2022digital,zhang2022artificial}. Advanced DT platforms increasingly incorporate multi-omics and environmental data for even greater personalization \cite{sun2023digital,mihai2022digital}.
Data integration presents significant technical challenges: harmonizing spatial, temporal, physiological, and procedural data streams requires sophisticated data fusion algorithms and interoperability standards \cite{das2022toward,chen2024networking}. Semantic data frameworks and standardized ontologies can further support cross-platform and multi-center DT applications \cite{sun2023digital,elayan2021digital,chen2024networking,saxby2023digital,lonsdale2022perioperative}.
Given the sensitivity of medical data, security and privacy are paramount. Regulatory compliance with frameworks such as GDPR and HIPAA is non-negotiable, and technical safeguards (encryption, access controls, federated learning) are essential to patient trust \cite{qin2022realizing,asciak2025digital,suraci2025migrate}. Privacy-preserving digital-twin encodings for OR video streams (semantic masks + monocular depth) illustrate practical approaches to data minimization while retaining analytic utility \cite{perez2025privacy}.
Cloud and edge computing infrastructures are crucial for supporting the real-time processing demands of surgical DTs, enabling rapid data analysis, predictive modeling, and feedback with minimal latency \cite{sun2023digital,elayan2021digital,tai2022digital,das2022toward}. Feedback loops close the digital-physical circuit, enabling bidirectional communication, adaptive control, and continuous learning \cite{elayan2021digital,das2022toward,asciak2025digital}.
For sustainable and ethical DT integration, the literature stresses the importance of standardization, transparent governance, and robust technical infrastructure \cite{sun2023digital,das2022toward,mihai2022digital,chen2024networking,saxby2023digital,lonsdale2022perioperative}. Addressing these domains will unlock the full transformative potential of DTs in both routine and emergency surgical environments.

\section{Decision-Making in Surgical Emergencies}

\subsection{Characteristics and Challenges of Surgical Emergencies}
Surgical emergencies present a unique intersection of unpredictability, physiological instability, and extreme time pressure. These situations, which include acute trauma, catastrophic hemorrhage, and sudden organ failure, demand rapid decision-making often under conditions of incomplete information and constrained resources. The inherent volatility of emergency cases—where patient status can deteriorate within seconds—exposes critical weaknesses in conventional decision support strategies, which frequently depend on manual data integration and retrospective reasoning \cite{chen2024networking},\cite{asciak2025digital},\cite{martinez2021machine}
A salient challenge in these contexts is the need for continuous assimilation and synthesis of heterogeneous data sources. Imaging modalities, intraoperative sensors, laboratory findings, and clinical records must be rapidly reconciled to present an actionable and temporally relevant clinical picture. Manual approaches to this task are not only time-consuming but also susceptible to cognitive overload and human error, especially as the complexity of both patient physiology and therapeutic options increases \cite{gong2023interactive},\cite{shi2022synergistic},\cite{suraci2025migrate}. These difficulties are magnified in distributed or remote settings, where bandwidth, specialist access, and equipment availability can be unpredictable \cite{laaki2019prototyping},\cite{das2022toward},\cite{suraci2025migrate}. Here, delays in communication or loss of data fidelity can have direct consequences for patient survival.

Furthermore, intraoperative complications such as unexpected bleeding or tissue injury often require immediate and coordinated action. In such scenarios, the absence of real-time, holistic information can force surgical teams to rely on experience or intuition—methods which, while valuable, are vulnerable to bias and cannot account for the dynamic interplay of all relevant variables \cite{annamraju2024digital},\cite{ding2024towards_automation}. It is thus apparent that effective decision support in surgical emergencies demands both a high degree of automation in data fusion and the ability to predict and simulate physiologic responses to potential interventions.

\subsection{Real-Time Decision-Making Requirements}
The imperative for rapid and accurate decision-making in surgical emergencies is underpinned by stringent requirements for system performance, particularly in terms of latency, reliability and integration. Any delay in hemorrhage control or airway management may result in irreversible damage or mortality, making it essential that decision support systems provide actionable guidance within clinically meaningful timeframes \cite{laaki2019prototyping, tai2022digital}.

Effective systems must accommodate the diversity and variability of clinical data. High-frequency physiological measurements, streaming video, preoperative imaging, and historical health data are typically recorded in incompatible formats and at varying time resolutions. The ability to harmonize these disparate sources and update risk assessments in real time is critical \cite{chen2024networking},\cite{gong2023interactive}. This process, however, is complicated by potential noise, gaps, and inaccuracies in the data, especially in the context of patient movement, emergency transportation, or equipment malfunction \cite{ding2024towards_automation},\cite{suraci2025migrate}.

Additionally, distributed care networks—such as those involving telemedicine or remote robotic surgery—require that decision support tools ensure secure, low-latency communication and preserve data integrity across geographic and institutional boundaries \cite{das2022toward},\cite{suraci2025migrate}. For these systems to be effective in emergencies, they must not only migrate digital twin models and clinical data seamlessly as patients move but also maintain real-time synchronization between local and remote teams.

\subsection{Roles of Digital Twins in Emergency Decision Support}
Digital twin technology has emerged as a transformative solution for many of the above challenges, offering a dynamic, patient-specific virtual representation that can be updated in real time and serve as a platform for both monitoring and simulation \cite{asciak2025digital},\cite{chen2024networking}. By continuously ingesting multimodal data—ranging from intraoperative sensor feeds to imaging and laboratory results—digital twins maintain an evolving model of the patient that reflects both current and historical physiologic states.

In the context of surgical emergencies, this capability enables several crucial forms of decision support. First, digital twins facilitate continuous real-time monitoring, providing clinicians with an integrated and updated view of patient status. This is particularly valuable during acute events, such as trauma resuscitation or intraoperative hemorrhage, where rapid physiologic changes can outpace the manual updating of information systems \cite{das2022toward},\cite{laaki2019prototyping}. Second, digital twins support simulation-based prediction. By leveraging computational models of tissue properties, hemodynamics, and procedural effects, these systems can anticipate the consequences of various interventions before they are undertaken, thus informing the selection and timing of critical maneuvers \cite{servin2024simulation},\cite{shi2022synergistic}.

For instance, Servin et~al.\ \cite{servin2024simulation},\cite{servin2025digital} have demonstrated the use of MRI-based digital twin models to simulate microwave ablation outcomes in liver emergencies, accounting for patient-specific fat distribution and tissue heterogeneity to optimize ablation margins. In cardiac emergencies, Gong et~al.\ \cite{gong2023interactive} describe an interactive digital twin platform for the human heart, allowing dynamic simulation of interventions and facilitating real-time intraoperative planning. When extended to augmented reality, such as in the work by Shi et~al.\ \cite{shi2022synergistic}, these platforms further enable intuitive, holographic visualization and navigation, greatly enhancing the surgeon’s situational awareness during critical procedures.

Importantly, digital twins also underpin remote and robotic surgical interventions. By maintaining a synchronized model between patient and remote operator, these systems facilitate the delivery of expert care across distances, compensating for network variability and enabling collaborative decision-making even in bandwidth-limited or high-latency environments \cite{laaki2019prototyping},\cite{das2022toward},\cite{suraci2025migrate}. The adaptive nature of these platforms allows them to dynamically adjust to intraoperative events and evolving patient states, providing a robust foundation for next-generation emergency care.

\subsection{Case Studies in Emergency Decision-Making}
Several recent studies exemplify the application and promise of digital twin-based decision support in surgical emergencies. Laaki et~al.\ \cite{laaki2019prototyping} report the development of a real-time digital twin platform for remote liver surgery, in which anatomical and physiological models are dynamically updated via mobile networks to guide transplantation and acute interventions across locations. This system enables both local and remote teams to share a unified, evolving model of the patient, allowing for collaborative monitoring and rapid adjustment of surgical strategies as emergencies unfold.

In the field of interventional oncology, Servin et~al.\ \cite{servin2024simulation}, \cite{servin2025digital} have shown that digital twins integrating spatially resolved, imaging-derived tissue properties can significantly improve ablation planning during acute liver emergencies. Their results indicate that accounting for tissue heterogeneity and fat content leads to more accurate predictions of ablation zones, reducing the risk of incomplete treatment and facilitating safer, more effective interventions.

Cardiac surgery has similarly benefited from digital twin approaches. Gong et~al.\ \cite{gong2023interactive} developed a high-performance, interactive heart model that enables surgeons to simulate and evaluate emergency procedures in real time. This platform supports both preoperative rehearsal and intraoperative guidance, improving the speed and precision of response to sudden cardiac events.

The use of digital twins in trauma and damage control surgery has been highlighted by Chen et~al.\ \cite{chen2024networking} and Das et~al.\ \cite{das2022toward}, who emphasize the need for rapid, distributed decision support capable of integrating real-time physiological, imaging, and contextual data. Their proposed architectures enable teams to synchronize emergency responses across multiple sites, share data instantaneously, and employ predictive analytics to prioritize and adapt interventions during mass casualty or multi-trauma events.

Finally, Shi et~al.\ \cite{shi2022synergistic} describe a synergistic digital twin and augmented reality navigation system for percutaneous liver tumor puncture under respiratory motion, achieving millimeter-scale targeting accuracy in animal models. Their approach, which dynamically adapts to predicted tissue motion, demonstrates the capacity for digital twin-based platforms to enhance procedural precision and safety in acute, motion-affected interventions.

\section{Mathematical Formulations}
\label{section4:Mathematical Formulations}

\subsection{Modeling Surgical Procedures}
The mathematical modeling of surgical procedures underpins the effectiveness of SDTs by providing the quantitative backbone necessary for simulation, prediction, and decision support. Within the literature, continuous, discrete, and hybrid mathematical frameworks are all utilized to model the rich dynamism and heterogeneity of the surgical environment, each with their own distinct advantages and application scenarios.

Continuous modeling strategies—predominantly grounded in systems of ordinary or partial differential equations—excel at capturing the physical and physiological aspects of surgery. These approaches enable the simulation of tissue deformation, thermal propagation, hemodynamic flows, and electrophysiological signals, as seen in patient-specific heart and liver models \cite{gong2023interactive},\cite{servin2023interactive},\cite{servin2024simulation},\cite{diniz2025digital}. For instance, the spatiotemporal evolution of thermal energy during microwave ablation can be described by the bioheat equation:

\begin{align}
\frac{\partial T}{\partial t} = \nabla \cdot (k \nabla T) + Q \label{eq:1.1}
\end{align}

where $T$ denotes temperature, $k$ is the thermal conductivity, and $Q$ the localized heat generation \cite{servin2023interactive},\cite{servin2024simulation}. Continuous finite element models have proven valuable in simulating the biomechanical response of musculoskeletal structures under surgical manipulation \cite{diniz2025digital}. However, these high-fidelity models often require extensive computational resources and comprehensive patient-specific parameterization, which may limit their practical deployment in real-time clinical workflows (as in Eq.~\eqref{eq:1.1})\cite{sun2023digital},\cite{diniz2025digital},\cite{ding2024unifying}.

Discrete modeling, in contrast, is well-suited to representing the event-driven nature of surgical workflows, procedural logic, and the sequencing of clinical tasks \cite{tai2022digital},\cite{qin2022realizing},\cite{asciak2025digital},\cite{hagmann2021digital},\cite{ding2024towards_automation},\cite{ding2024unifying}. Discrete-event simulations, Petri nets, and agent-based models abstract the surgical process as a sequence of states and transitions, facilitating granular analysis of workflow efficiency, error propagation, and robotic tool choreography \cite{tai2022digital},\cite{hagmann2021digital}. Such models are integral to the development of context-aware surgical training systems and process monitoring platforms \cite{asciak2025digital},\cite{hagmann2021digital} as well as for encoding adaptive robotic control policies in teleoperated environments \cite{das2022toward},\cite{wang2025digital},\cite{bonne2022digital}. However, their limited capacity to capture continuous biophysical dynamics can constrain their utility in applications where physiological fidelity is essential.

Recognizing the limitations inherent in purely continuous or discrete approaches, hybrid modeling strategies have emerged as a state-of-the-art solution for surgical digital twins. These frameworks integrate continuous dynamic models with discrete state machines, enabling the simultaneous tracking of physiological changes and procedural events \cite{tai2022digital},\cite{annamraju2024digital},\cite{hagmann2021digital},\cite{ding2024unifying}. For example, hybrid automata have been deployed to synchronize real-time updates of tissue deformation with the sequencing of surgical tool actions \cite{annamraju2024digital},\cite{hagmann2021digital}. This integration allows digital twins to provide context-sensitive support, adapting to abrupt workflow transitions and physiological perturbations alike—a critical requirement in both routine and emergent surgical scenarios \cite{tai2022digital},\cite{asciak2025digital},\cite{annamraju2024digital}.

\subsection{Mutual Information and Information-Theoretic Approaches}
With the proliferation of multi-modal surgical data, information-theoretic concepts have become indispensable for quantifying uncertainty, relevance, and redundancy across heterogeneous information streams. Mutual information $I(X;Y)$ has been adopted as a rigorous metric for selecting the most informative sensors, imaging modalities, or clinical parameters, thereby enhancing the efficiency and robustness of digital twin–enabled decision support \cite{sun2023digital},\cite{elayan2021digital},\cite{das2022toward},\cite{qin2022realizing},\cite{asciak2025digital},\cite{suraci2025migrate},\cite{mihai2022digital},\cite{chen2024networking},\cite{ding2024towards_automation}. This is formally captured as Eq. \eqref{eq:1.2}:

\begin{equation}
I(X;Y) = \sum_{x,y} p(x,y) \log \frac{p(x,y)}{p(x)p(y)}  
\label{eq:1.2}
\end{equation}

where $p(x,y)$ is the joint probability distribution, and $p(x), p(y)$ are the marginals. In the surgical context, mutual information guides adaptive data fusion, allowing the digital twin to prioritize data channels whose real-time updates maximally reduce uncertainty regarding patient state or surgical outcome (as in Eq.~\eqref{eq:1.2}) \cite{qin2022realizing},\cite{asciak2025digital},\cite{chen2024networking},\cite{ding2024towards_automation}.

A substantial body of work has harnessed Markov decision processes (MDPs) as the mathematical foundation for optimizing sequential clinical decisions under uncertainty—a pervasive feature in both planned and emergent surgical care. In an MDP, the system is characterized by a set of states $S$, a set of actions $A$, a transition probability function $P(s'|s,a)$, a reward function $R(s,a)$, and a discount factor $\gamma$ \cite{martinez2021machine},\cite{sapkota2024machine},\cite{ding2024unifying}. The objective is to identify a policy $\pi$ that maximizes the expected cumulative reward as Eq. \eqref{eq:1.3}:

\begin{align}
V(s) = \mathbb{E} \bigg[\sum_{t=0}^{\infty} \gamma^t R(s_t,a_t) \,\bigg|\, s_0 = s \bigg]  
\label{eq:1.3}
\end{align}

Advanced variants, such as partially observable MDPs (POMDPs), extend the model to situations where the true state is not fully observable, and decisions must be based on a probability distribution (belief state) over possible states \cite{servin2024simulation},\cite{servin2025digital},\cite{ding2024unifying}. Formally, a POMDP is defined by the tuple $(S, A, O, P, Z, R, \gamma)$, where $Z(o|s',a)$ represents the observation probability, as shown in Eq.~\eqref{eq:1.3}. 

Hierarchical and distributed MDPs facilitate multi-level modeling, spanning decisions from tool actuation up to strategic workflow management \cite{tai2022digital},\cite{asciak2025digital},\cite{annamraju2024digital},\cite{ding2024towards_automation},\cite{ding2024unifying}. Risk-sensitive and constrained MDPs have been deployed in trauma and high-stakes environments to balance anticipated benefits against catastrophic risk \cite{martinez2021machine, sapkota2024machine, ding2024unifying}.

\begin{figure}
    \centering
    \includegraphics[width=\linewidth]{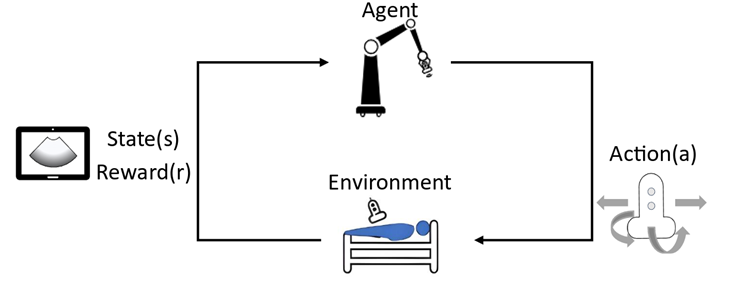}
    \caption{An illustration of agent-environment interaction in RL under the context of surgical robots \cite{qian2023deep}.}
    \label{fig:Agent-Environment Interaction}
\end{figure}

\subsection{Optimization under Uncertainty and Risk Constraints}
Optimization methods are foundational to digital twin–enabled surgical planning, resource allocation, and adaptive intraoperative management. Stochastic optimization extends classical formulations to explicitly incorporate probabilistic models of parameter variability and measurement noise:

\begin{align}
\min_{x \in X} \; \mathbb{E}_{\xi}\big[ f(x, \xi) \big]  \label{eq:1.4}
\end{align}

where $x$ denotes the decision variables, $\xi$ the random variables capturing uncertainty, and $f$ the performance metric (e.g., outcome, utility) (as in Eq.~\eqref{eq:1.4}) \cite{tai2022digital,martinez2021machine,servin2025digital,sapkota2024machine,ding2024unifying}.

Risk-sensitive optimization introduces loss functions—such as Value-at-Risk (VaR) and Conditional Value-at-Risk (CVaR)—to penalize strategies with high exposure to catastrophic complications, thereby generating plans tailored to a patient’s risk profile \cite{sun2023digital},\cite{tai2022digital},\cite{asciak2025digital},\cite{servin2023interactive},\cite{diniz2025digital}. Robust optimization further seeks solutions that perform well under the worst-case realization of uncertainties:

\begin{align}
\min_{x \in X} \max_{\xi \in \Xi} f(x, \xi) \label{eq:1.5}
\end{align}

with $\Xi$ representing the uncertainty set (as in Eq.~\eqref{eq:1.5}) \cite{das2022toward},\cite{mihai2022digital},\cite{khan2022digital},\cite{ding2024unifying}. Recent advances couple optimization with online machine learning, enabling digital twins to update decision policies in real time as new data are assimilated, thus supporting a continuous learning paradigm.

\begin{table*}[htbp]
\centering
\caption{Comparative Summary of Mathematical Modeling Approaches in Surgical Digital Twins}
\label{tab:math_approaches}
\renewcommand{\arraystretch}{1.2}
\begin{tabularx}{\textwidth}{@{}p{0.15\textwidth} X X X X@{}}
\toprule
\textbf{Modeling Approach} & \textbf{Mathematical Tools} & \textbf{Primary Applications} & \textbf{Strengths} & \textbf{Weaknesses} \\
\midrule
Continuous & PDEs, ODEs, Finite Element Methods & Biomechanical modeling, ablation, cardiac simulation & High physiological fidelity & Computational cost, model calibration \\
Discrete & DES, Petri Nets, Agent-Based Models & Workflow, tool sequence, training simulators & Captures logic, efficient for process modeling & Limited biophysical realism \\
Hybrid & Hybrid Automata, Coupled Simulation & Adaptive DTs, context-aware guidance & Flexibility, context sensitivity & Model integration complexity \\
Information-Theoretic & Mutual Information, Entropy, Data Fusion & Sensor selection, communication optimization & Quantifies data value, manages uncertainty & Probability estimation challenges \\
MDP (and variants) & States, Actions, Rewards, POMDPs, Hierarchies & Sequential and adaptive decision-making & Handles uncertainty, enables learning & Computationally intensive in large spaces \\
Optimization & Stochastic, Robust, Risk-sensitive Programming & Resource allocation, risk-aware planning & Manages uncertainty, tail risk, continuous updating & Solution conservatism, computation time \\
\bottomrule
\end{tabularx}
\end{table*}

\section{Connections with DEVS (Discrete-Event System Specification)}
\label{section5:Connections with DEVS}

Surgical care is full of discrete events—incision, imaging updates, alarms, decisions, hand-offs—arriving at irregular times and often under uncertainty. The Discrete-Event System Specification (DEVS) is a modeling framework built precisely for such event-driven, asynchronous systems. It decomposes a procedure into modular components that react to events, pass messages, and synchronize with real-time data streams, making it a natural backbone for surgical digital twins that must “keep up” with the operating room \cite{zeigler2000theory}. In contrast to continuous-only models, DEVS handles rare events and branching workflows without losing timing fidelity, and it composes cleanly with physics simulators, AR/VR guidance, and robot or non-robot interfaces \cite{sun2023digital},\cite{elayan2021digital},\cite{das2022toward},\cite{asciak2025digital}.

In practice, DEVS is used to (i) encode workflow logic and decision points, (ii) co-simulate with biophysical models (e.g., ablation or biomechanics) for prediction, and (iii) integrate data from sensors, imaging, and networks so the twin updates when the patient or environment changes \cite{tai2022digital},\cite{gong2023interactive},\cite{servin2023interactive},\cite{servin2024simulation},\cite{xie2024tiodt},\cite{suraci2025migrate}. This section explains how DEVS maps to surgical tasks, couples with the rest of the SDT stack, and compares with other formalisms like Agent-Based Modeling (ABM), System Dynamics (SD), Petri Nets, and BPMN/UML \cite{sun2023digital},\cite{mihai2022digital},\cite{shi2022synergistic},\cite{zeigler2000theory}.

\subsection{Overview of the DEVS Modeling Framework}
DEVS is modular and hierarchical: systems are represented as \emph{atomic models} (react to inputs, schedule internal transitions, emit outputs) composed into \emph{coupled models} (networks of interacting components) \cite{zeigler2000theory}. For surgical digital twins, this mirrors peri-op intake, anesthesia, imaging, intra-op decisions, and post-op care as interacting event-driven subsystems with explicit timing. The decomposition supports clear interfaces, real-time accuracy, and rapid response to unexpected events such as hemorrhage or device faults \cite{asciak2025digital},\cite{annamraju2024digital},\cite{hagmann2021digital}. DEVS also provides a unified way to fuse heterogeneous IoT, robot, and EHR streams into a single evolving model of the case \cite{elayan2021digital},\cite{das2022toward},\cite{mihai2022digital}, enabling synchronization with reality in procedure and emergency settings \cite{asciak2025digital},\cite{suraci2025migrate}.

\subsection{DEVS Applications in Surgical Simulations}
Event-centric simulators align with how clinicians think about workflows and decision points. DEVS architectures have been used to capture the stochastic, sequential nature of surgical actions and tool–tissue interactions in IoMT-enabled simulations \cite{tai2022digital}, and to coordinate perception, manipulation, and feedback in minimally invasive scenarios \cite{xie2024tiodt}. Because emergency surgery hinges on rare, time-critical events, DEVS enables exploration of alternatives, modeling of error cascades, and recovery strategies—critical for training and protocol optimization \cite{asciak2025digital},\cite{annamraju2024digital},\cite{hagmann2021digital},\cite{shi2022synergistic}. Event-driven updates can also keep AR overlays synchronized with the evolving operative field without imposing tight control loops \cite{shi2022synergistic}.

\subsection{Integrating DEVS with Surgical Digital Twins}
Within an SDT, DEVS can serve as the orchestration layer: it manages discrete updates triggered by patient state changes, sensor events, and OR signals, while co-simulation engines advance continuous biophysical models (e.g., FEA, bioheat) between events \cite{gong2023interactive},\cite{servin2023interactive},\cite{servin2024simulation},\cite{annamraju2024digital}. Middleware enables distributed DEVS components to remain synchronized across edge/cloud setups and multi-site teams, supporting telesurgery and remote collaboration \cite{das2022toward},\cite{xie2024tiodt},\cite{suraci2025migrate}. Case studies include interactive cardiac twins that respond to device activations and arrhythmia events in real time \cite{gong2023interactive}, and scene-representation approaches for event-driven perception and robust automation/decision support in the OR \cite{ding2024towards_automation},\cite{ding2024unifying}.

\subsection{Comparison with Other Modeling Frameworks}
While DEVS excels at discrete-event modeling, it is complemented by other frameworks. A high-level comparison is summarized in Table~\ref{tab:framework_comparison}.

\begin{table*}[htbp]
\centering
\caption{Comparison of Modeling Frameworks for Surgical and Healthcare Digital Twins}
\label{tab:framework_comparison}
\renewcommand{\arraystretch}{1.2}
\begin{tabularx}{\textwidth}{@{}p{0.16\textwidth} X X X X X@{}}
\toprule
\textbf{Aspect / Method} & \textbf{DEVS} & \textbf{ABM} & \textbf{SD} & \textbf{Petri Nets} & \textbf{BPMN/UML} \\
\midrule
Core Focus & Discrete event, modular, hierarchical simulation & Emergent, agent-level interactions & Aggregate, continuous feedback systems & Concurrency, workflow, resource flow & Process visualization \& documentation \\
Best Use Cases & Event-driven surgical workflows, real-time DT & Decentralized, adaptive multi-agent systems & Policy, long-term trends, chronic disease & Workflow bottleneck/deadlock analysis & Communication, process documentation \\
Strengths & Accurate event timing, extensible, real-time capable & Emergent behavior, heterogeneity & High-level abstraction, feedback loops & Formal verification tools & Clear visual logic, easy adoption \\
Weaknesses & Complex for very large agent populations & Inefficient for tightly sequenced tasks & Cannot model rare events well & Limited hierarchy, scalability issues & Limited simulation capability \\
Granularity & Fine-grained discrete events & Individual agent level & Aggregate/system level & Intermediate (resources/workflows) & High-level process steps \\
Extensibility & High, supports hierarchy & Moderate, modular agents & Low, not modular & Moderate, limited hierarchy & Low, mainly documentation \\
\bottomrule
\end{tabularx}
\end{table*}

\section{Robotic Surgical Systems and Digital Twin Integration}
\label{section6:Robotic Surgical System and Digital Twin Integration}

\subsection{Overview of Robotic Surgical Systems}
Robotic-assisted surgery has rapidly advanced minimally invasive interventions, offering increased dexterity, precision, and visualization compared to conventional techniques. The da Vinci Surgical System is the most widely adopted platform, featuring a surgeon console, multi-armed patient-side manipulators, and advanced imaging that support intricate procedures with minimal incisions \cite{hagmann2021digital}, \cite{asciak2025digital}, \cite{annamraju2024digital}. Other platforms, such as the DLR MiroSurge and Intuitive’s Ion, broaden the landscape across urology, gynecology, thoracic, and cardiac surgery \cite{hagmann2021digital}, \cite{diniz2025digital}.

Core capabilities—high-fidelity motion control, tremor filtration, stereoscopic visualization, and ergonomic design—enable complex tasks (suturing, dissection, knot-tying) with sub-millimeter precision \cite{hagmann2021digital}, \cite{diniz2025digital}. Newer systems incorporate haptics, instrument tracking, and semi-autonomous features, paving the way for tighter synergy with cyber-physical and computational systems \cite{hagmann2021digital}, \cite{annamraju2024digital}. Despite these advances, outcomes still depend heavily on operator skill and situational awareness, motivating enhanced decision support \cite{asciak2025digital}, \cite{hagmann2021digital}.

\begin{figure}
    \centering
    \includegraphics[width=\linewidth]{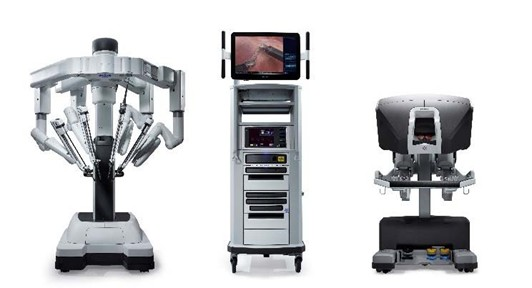}
    \caption{Components of the da Vinci Surgical System: patient-side cart, vision cart, surgeon console \cite{avgousti2020robotic}.}
    \label{fig:da Vinci Surgical System}
\end{figure}

\subsection{Digital Twin Integration with Robotic Surgical Systems}
Integrating digital twins (DTs) with robotic platforms synchronizes a real-time, patient-specific virtual model with the operative environment. This depends on robust interfaces and middleware for bidirectional exchange among the DT, robot, and perioperative information systems, using standardized protocols for interoperability and low-latency communication \cite{asciak2025digital}, \cite{hagmann2021digital}, \cite{annamraju2024digital}, \cite{ding2024towards_automation}, \cite{das2022toward}, \cite{wang2025digital}, \cite{bonne2022digital}.

A closed-loop control/feedback cycle is central: the DT ingests intraoperative data (force, pose/kinematics, endoscopic video) and updates its virtual state to simulate tissue deformation, tool–tissue interaction, and physiological responses—anticipating adverse events and suggesting corrective actions before they materialize \cite{hagmann2021digital}, \cite{ding2024towards_automation}, \cite{ding2024unifying}, \cite{tai2022digital}, \cite{annamraju2024digital}]. In telesurgery, DTs help synchronize remote and local contexts, compensating for delay/jitter so teams share an up-to-date situational picture \cite{wang2025digital}, \cite{bonne2022digital}.

\begin{table*}[htbp]
\centering
\caption{Robotic platforms and DT integrations: features, strengths, and gaps.}
\label{tab:robotic_dt_integration}
\renewcommand{\arraystretch}{1.15}
\begin{tabularx}{\textwidth}{@{}p{0.15\textwidth} p{0.25\textwidth} p{0.15\textwidth} p{0.15\textwidth} p{0.2\textwidth}@{}}
\toprule
\textbf{Platform/ line of work} & \textbf{Integration features (DT $\leftrightarrow$ robot)} & \textbf{Strengths} & \textbf{Limitations} & \textbf{Gaps/ next steps} \\
\midrule
da Vinci (research integrations) \cite{asciak2025digital}, \cite{ding2024towards_automation}, \cite{ding2024unifying} &
DT synchronized with endoscopic video and kinematics; scene-level geometric representations for perception/ decision support &
Mature ecosystem; rich sensing; active benchmarks &
Proprietary interfaces; limited open middleware; tight real-time constraints &
Standardized APIs; robust vision+kinematics fusion; clinical V\&V of DT-assisted decisions \\
\addlinespace
DLR MiroSurge (training/ assistance) \cite{hagmann2021digital} &
Context-aware assistance, virtual fixtures, haptics; DT for training/guidance &
High-fidelity haptics; modular research platform &
Primarily training/ prototyping; limited live-OR evidence &
Translate to clinical workflows; user-in-the-loop safety studies; comparative trials vs.\ non-DT training \\
\addlinespace
TIODT platform for MIS \cite{xie2024tiodt} &
Touch-free DT interaction; event-driven coordination of imaging/tools &
Human-factors focus; reduced cognitive load &
Early-stage; narrow task scope; device dependencies &
Broaden task coverage; latency/robustness studies; IRB-approved evaluations \\
\addlinespace
Telesurgery under intermittent comms \cite{wang2025digital} &
DT maintains synchronized state; buffers/compensates for link intermittency; network-aware control &
Designed for variable networks; demonstrates feasibility &
Delay/jitter limit tight control; limited clinical validation &
QoS-aware DT scheduling; safety envelopes under delay; multi-site trials \\
\addlinespace
QoS-aware DT framework for telesurgery \cite{bonne2022digital} &
DT explicitly models network QoS; adaptation under varying bandwidth/latency &
Formalizes network effects; CASE-validated framework &
Prototype-level; limited modality coverage &
Joint DT–network co-design; standard KPIs and benchmarks \\
\addlinespace
IoRT collaborative DT frameworks \cite{das2022toward}, \cite{qamsane2019unified} &
Middleware orchestrates devices/ sensors; DT threads across OR endpoints &
Scales to many devices; cyber-physical design fit &
Interoperability uneven; security posture varies &
Common data models/APIs; conformance testing; security hardening \\
\addlinespace
Perception-centric DTs for autonomy \cite{ding2024towards_automation}, \cite{ding2024unifying} &
DT as scene representation; foundation-model perception feeds planning/control &
Robust perception under variation; path to autonomy &
Dataset bias; OR compute cost &
Real-time guarantees; domain-shift handling; surgeon oversight integration \\
\addlinespace
IoMT/edge-sensor integrated DTs \cite{tai2022digital}, \cite{suraci2025migrate}, \cite{chen2024networking}, \cite{bulej2021managing} &
Event-driven ingestion (vision, trackers, physio); DT updates for rehearsal/guidance &
Fits multimodal OR data; supports what-if rehearsal &
Often offline/semi-online; perf bounds under-reported &
Edge/cloud partitioning; latency budgets and SLAs \\
\bottomrule
\end{tabularx}
\end{table*}

\subsection{Case Studies: da Vinci System Applications}
DLR MiroSurge applies a DT framework for training/planning with real-time haptics and contextual assistance \cite{hagmann2021digital}. In multi-center simulations, DTs have been coupled to da Vinci simulators for collaborative rehearsal, integrating anatomical models and instrument trajectories to assess skills and optimize workflows \cite{tai2022digital}, \cite{hagmann2021digital}. Real-time intraoperative guidance updates patient-specific biomechanical models from da Vinci sensors and endoscopic imaging, enabling what-if testing and predictive analytics for critical decisions \cite{asciak2025digital}, \cite{ding2024unifying}, \cite{annamraju2024digital}, \cite{ding2024towards_automation}. DTs can detect deviations from expected tissue response or tool trajectory and issue alerts/suggestions to improve safety and efficiency [20, \cite{ding2024unifying}.

\subsection{Challenges in Robotic Integration}
Key barriers include latency in data transfer/processing that undermines synchronization—especially for telesurgery or cloud-backed computation \cite{das2022toward}, \cite{wang2025digital}, \cite{bonne2022digital}—and interoperability problems due to proprietary interfaces and heterogeneous clinical data standards \cite{asciak2025digital}, \cite{hagmann2021digital}, \cite{chen2024networking}, \cite{ding2024unifying}. Maintaining DT fidelity amid noisy, incomplete, or delayed inputs requires advanced filtering and predictive models \cite{asciak2025digital}, \cite{annamraju2024digital}, \cite{ding2024towards_automation}, \cite{ding2024unifying}. Security and privacy concerns mandate strong protocols and access controls \cite{chen2024networking}, \cite{ding2024unifying}. Finally, translating DT recommendations into robotic actions with appropriate operator oversight requires rigorous validation and regulatory clearance, complicating clinical adoption \cite{ahmed2020potential}, \cite{asciak2025digital}, \cite{martinez2021machine}.

\section{Communication and Simulation Tools for Digital Twin}
\label{section7:Communication and Simulation Tools}

\begin{figure}
    \centering
    \includegraphics[width=0.9\linewidth]{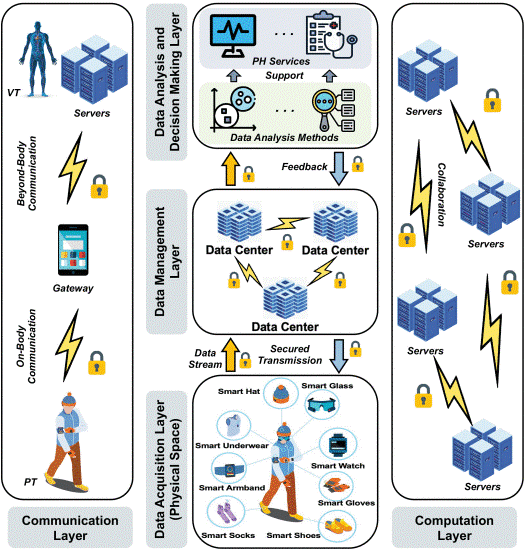}
    \caption{Networking architecture of the Human Digital Twin\cite{chen2024networking}.}
    \label{fig:Network Architecture of HDT}
\end{figure}

\subsection{Communication Tools in DT (5G/6G, IoT, Edge)}
Robust communication frameworks are essential to keep DTs synchronized with the physical world under clinical time constraints. 5G delivers ultra-reliable low-latency links (down to $\sim 1\,\mathrm{ms}$ in ideal cases) with network slicing, while 6G targets sub-ms latency and AI-native orchestration \cite{laaki2019prototyping},\cite{wang2025digital},\cite{suraci2025migrate},\cite{chen2024networking},\cite{bonne2022digital}. IoT devices provide continuous, multimodal data feeds; edge computing processes data close to the source to reduce delay and enhance privacy, often outperforming cloud-only pipelines by $20\text{--}30\%$ latency reductions in medical DT migration studies \cite{suraci2025migrate},\cite{chen2024networking}. Hybrid edge–cloud designs offload heavy simulation (e.g., FEA/bioheat) while keeping guidance loops local \cite{wang2025digital},\cite{suraci2025migrate},\cite{yin2019realtime},\cite{bulej2021managing}.

\begin{table*}[htbp]
\centering
\caption{Comparison of communication technologies for healthcare/surgical DTs (indicative latencies).}
\label{tab:dt_comm_compare}
\renewcommand{\arraystretch}{1.15}
\begin{tabularx}{\textwidth}{@{}p{0.19\textwidth} p{0.12\textwidth} X X X@{}}
\toprule
\textbf{Technology} & \textbf{Latency (ms)} & \textbf{Main advantage} & \textbf{Main limitation} & \textbf{Best-fit DT uses} \\
\midrule
5G (SA / slicing / private) \cite{laaki2019prototyping},\cite{wang2025digital},\cite{bonne2022digital} & $\sim 1\text{--}10$ (ideal); tens E2E & Low latency, high reliability/bandwidth; slicing & Standards still evolving; deployment scope & In-hospital DT loops; telesurgery pilots; high-rate vision/telemetry \\
6G (projected) \cite{suraci2025migrate},\cite{chen2024networking} & $<1$ (target) & AI-native networking, tight edge integration & Early-stage research & Future ultra-reliable OR networking; DT migration/ orchestration \\
IoT + Edge (on-prem/ OR edge) \cite{elayan2021digital},\cite{suraci2025migrate},\cite{chen2024networking} & $\sim 1\text{--}50$ (pipeline-dependent) & Local processing $\Rightarrow$ lower latency; privacy/ resilience & Heterogeneous protocols; interoperability & Real-time sensing/ fusion; AR overlays; local analytics \\
Cloud-only \cite{suraci2025migrate} & $\sim 50\text{--}500+$ & Centralized scale-out resources & Higher latency; governance constraints & Non-urgent analytics; post-hoc model training \\
Hybrid edge–cloud \cite{wang2025digital},\cite{suraci2025migrate},\cite{yin2019realtime},\cite{bulej2021managing} & $\sim 10\text{--}100+$ (path-dependent) & Elastic compute; offload heavy sim to cloud & Partitioning complexity; failure modes under churn & Mixed workloads (remote FEA/ bioheat; local guidance/ control) \\
\bottomrule
\end{tabularx}
\end{table*}

\paragraph{Challenges and mitigations.}
Network latency/jitter can destabilize surgical DT loops, especially over wireless or remote links; mitigations include multi-path transport, adaptive buffering, and AI-based QoS prediction \cite{laaki2019prototyping},\cite{wang2025digital},\cite{bonne2022digital}. Interoperability across heterogeneous IoT stacks remains difficult, motivating open data models and middleware. Security/privacy require end-to-end encryption, secure edge processing, and federated learning for sensitive data \cite{suraci2025migrate},\cite{chen2024networking}.

\subsection{Simulation Tools in DT (VR/AR)}
VR/AR platforms make DTs interactive and immersive for rehearsal, guidance, and training. VR systems (e.g., “VR Isle Academy”) emphasize full-procedure rehearsal and skills analytics; AR overlays (e.g., holographic puncture guidance) project landmarks/trajectories onto the operative field for live assistance \cite{shu2023twins},\cite{filippidis2024vr},\cite{shi2022synergistic},\cite{aloqaily2023digital},\cite{klimo2023digital}. Detailed 3D human models derived from imaging can be deployed in both VR and AR, subject to software/hardware support \cite{aloqaily2023digital},\cite{klimo2023digital}. DT–AR/VR convergence enables real-time updates, multi-user collaboration, and context-aware decision support.

\begin{table*}[htbp]
\centering
\caption{Summary of VR/AR-based simulation platforms in DT systems.}
\label{tab:vr_ar_summary}
\renewcommand{\arraystretch}{1.15}
\begin{tabularx}{\textwidth}{@{}p{0.2\textwidth} p{0.08\textwidth} X X X@{}}
\toprule
\textbf{Simulation tool / ref.} & \textbf{Modality} & \textbf{Primary use case} & \textbf{Strength / advantage} & \textbf{Limitation} \\
\midrule
VR Isle Academy \cite{filippidis2024vr} & VR & Robotic skill training & Immersive realism; scenario diversity & Hardware cost; limited haptics \\
3D Human Body Models \cite{klimo2023digital} & VR \& AR & Anatomy simulation; preop planning & High anatomical fidelity & Needs patient-specific data/pipelines \\
DT + Holographic AR \cite{shi2022synergistic} & AR & Real-time intraop guidance & Contextual overlays; improved targeting accuracy & Field of view; hardware complexity \\
Immersive DT platforms \cite{aloqaily2023digital} & VR \& AR & Education; visualization; collaboration & Multi-user support; adaptability & Interoperability, live data integration \\
Twin-S (skull base) \cite{shu2023twins} & VR & Surgical rehearsal; risk assessment & Real-time updates; depth cues; sub-mm accuracy (phantoms) & Tracking line-of-sight limits; clinical validation pending \\
\bottomrule
\end{tabularx}
\end{table*}

\subsection{User Interface (UI) and Human Factors}

User interface (UI) design and human factors are central to the clinical adoption and safety of digital twin (DT) technologies in surgery and broader healthcare contexts. As DT systems increasingly integrate multimodal data streams—spanning real-time patient monitoring, AI-based recommendations, and advanced visualization—the complexity of surgeon–system interaction poses new challenges for usability, cognitive load, and workflow integration \cite{coorey2022health,badash2016innovations,huang2019telemedicine,gross1998computer,chen2007human}.

Contemporary research highlights the importance of adaptive, context-aware UIs capable of dynamically prioritizing and filtering information according to clinical urgency, user expertise, and procedural phase. Coorey et al.\ emphasize that next-generation DT user interfaces must support seamless multi-device interoperability—from workstations and tablets to immersive AR/VR and wearables—to maintain situational awareness and workflow continuity across perioperative environments \cite{coorey2022health}.

A critical consideration is the balance between information richness and cognitive load. Overly dense visualizations risk overwhelming clinicians during high-stakes scenarios, whereas insufficient detail may compromise decision quality \cite{huang2019telemedicine,chen2007human}. Recent advances in HCI for DTs recommend layered/adaptive displays that reveal additional data on demand and according to the user’s real-time context, complemented by multimodal I/O (gesture, touch, voice) to enable hands-free, sterile operation \cite{huang2019telemedicine,chen2007human}.

The usability of DT interfaces is further influenced by the adoption of immersive technologies. VR/AR-based interfaces support high-fidelity 3D visualization of patient anatomy, procedural simulations, and intraoperative navigation, with reported improvements in spatial understanding and error reduction \cite{klimo2023digital,chen2007human,coorey2022health}. However, practical barriers persist: hardware ergonomics, display latency, and integration with existing hospital IT systems remain unresolved and can limit user acceptance \cite{chen2007human,rajkomar2019machine}.

Human factors studies reinforce the value of participatory, user-centered design. Iterative co-design with surgeons, nurses, and staff can surface workflow pain points and lead to interfaces that are both intuitive and clinically grounded \cite{bruynseels2018digital,huang2019telemedicine}. Objective metrics (task time, error rate, eye tracking) are increasingly paired with subjective measures (e.g., NASA-TLX workload, trust, satisfaction) \cite{coorey2022health,huang2019telemedicine,chen2007human}.

Interface personalization and accessibility are also essential. DT systems should adapt UI complexity, color schemes, layouts, and control modalities to individual user needs—including visual/ergonomic considerations and cognitive fatigue during prolonged procedures \cite{kamel2021digital,huang2019telemedicine}.

Despite progress, key barriers persist: cognitive overload, inconsistent interaction metaphors, and insufficient standardization across DT platforms \cite{chen2007human,huang2019telemedicine}. Surveys also note that inadequate training and resistance to novel UI paradigms slow adoption even where benefits are demonstrable. Addressing these issues will require ongoing collaboration among HCI researchers, clinicians, engineers, and vendors, and a shift toward standardized, validated, and transparent usability evaluation protocols \cite{badash2016innovations,huang2019telemedicine,gross1998computer}.

\begin{table}[htbp]
\centering
\caption{Summary of key human factors and UI design strategies for DT systems.}
\label{tab:ui_hf_strategies_12_1}
\renewcommand{\arraystretch}{1.15}
\begin{tabularx}{\columnwidth}{@{}p{0.1\textwidth} X X X@{}}
\toprule
\textbf{Design strategy} & \textbf{Example implementation} & \textbf{Reported benefit} & \textbf{Limitation / challenge} \\
\midrule
Context-aware, adaptive UI \cite{huang2019telemedicine,chen2007human,coorey2022health} & Dynamic alerting; layered/on-demand info & Reduced cognitive load; improved safety & Risk of missed info if poorly tuned \\
Multimodal interaction \cite{gross1998computer,chen2007human} & Gesture/voice for sterile control & Hands-free use; improved ergonomics & Device compatibility; noise robustness \\
Immersive 3D visualization \cite{klimo2023digital,chen2007human,coorey2022health} & VR/AR overlays; anatomy simulations & Enhanced spatial understanding & Hardware cost; workflow fit; latency \\
User-centered co-design \cite{bruynseels2018digital,huang2019telemedicine,badash2016innovations} & Participatory workshops; iterative prototyping & Better workflow fit; higher acceptance & Time/resource intensive \\
Personalization \& accessibility \cite{huang2019telemedicine,kamel2021digital,chen2007human} & Adjustable color/layout/text size; fatigue-aware modes & Broader usability; reduced fatigue & Integration into commercial systems \\
\bottomrule
\end{tabularx}
\end{table}

\section{Computational and Algorithmic Aspects}

\subsection{Real-Time Computational Methods}
Real-time computational capability forms the backbone of advanced digital twin (DT) systems, ensuring timely synchronization between the physical entity and its digital replica. To meet clinical latency/throughput targets, implementations combine edge computing \cite{suraci2025migrate},\cite{chen2024networking},\cite{bulej2021managing}, hybrid cloud--edge data processing \cite{qi2018digital},\cite{wang2021digital}, parallelization/modularization \cite{yin2019realtime}, and GPU-accelerated simulation \cite{servin2023interactive},\cite{servin2024simulation},\cite{servin2025digital},\cite{klimo2023digital}. 

Architecturally, big-data pipelines must be paired with low-latency paths to balance accuracy and responsiveness \cite{qi2018digital}. In medical DTs for teleoperation, optimized encoding and multi-threaded software on robust networks can keep end-to-end delay under $100\,\mathrm{ms}$ \cite{laaki2019prototyping}. Plant-wide monitoring frameworks show that modular, parallel data streams sustain real-time performance \cite{yin2019realtime}. On the hardware/topology side, edge servers reduce round-trip times by offloading time-critical tasks from the cloud \cite{suraci2025migrate},\cite{chen2024networking},\cite{bulej2021managing}; strategic latency management (partitioning, local caching) can cut end-to-end delay by up to $50\%$ under high load \cite{bulej2021managing}. 

Efficient algorithms are equally crucial: fast incremental learning and state estimation \cite{fuller2020digital}, and virtual--real fusion learning \cite{wang2021digital} have improved inference speed/accuracy in DT-like settings. In image-guided procedures, GPU-accelerated grids have reduced ablation simulations to clinically usable windows (seconds to minutes) \cite{servin2023interactive},\cite{servin2024simulation},\cite{servin2025digital},\cite{klimo2023digital}. With motion compensation and AR guidance, targeting errors of $2.18$--$2.79\,\mathrm{mm}$ have been reported \cite{shi2022synergistic}.

\subsection{Data-Driven and AI-Based Approaches}
Data-driven and AI-enabled methods synthesize heterogeneous data streams to produce actionable guidance. Compared with purely model-based approaches, big data analytics improves adaptation to variability at the cost of higher compute demand \cite{qi2018digital}. For perioperative logistics, ML models (e.g., bagged trees) reduce $\mathrm{RMSE}$ by $10$--$20\%$ for surgery-duration prediction, supporting OR scheduling and resource allocation \cite{martinez2021machine},\cite{sapkota2024machine}. 

For sensor fusion in DTs, surveys highlight Bayesian inference, Kalman filtering, and deep learning, where the trade-off is between accuracy, sample efficiency, and interpretability \cite{blasch2021ml}. Bibliometrics on fault detection show a shift toward AI-based diagnosis for speed and scalability \cite{alauddin2018bibliometric}, while broader healthcare reviews document AI's predictive utility for disease risk assessment \cite{rong2020ai}. Big data integration across operational streams informs sustainability-style monitoring (transferable to holistic patient monitoring) \cite{tseng2020future}. In oncology, multi-modal image fusion with AI has shown faster, more accurate real-time diagnosis than single-modality baselines, shrinking time-to-diagnosis from hours to minutes \cite{bagheriye2025advancements}.

\subsection{Scalability and Computational Efficiency}
Scaling DTs demands distributed compute, storage, and networking. Hybrid cloud--edge architectures enable distributed processing and alleviate central bottlenecks \cite{qi2018digital}, while virtual--real fusion learning scales to hundreds of sensor streams without proportional latency growth via efficient resource scheduling and data prioritization \cite{wang2021digital}. 

Edge-based AI strategies (dynamic resource allocation, federated learning, auto-scaling containers) sustain real-time performance under workload variability \cite{santoso2024maximizing}. Unified DT frameworks that prioritize critical streams and use distributed event-driven designs achieve sub-second responsiveness at scale \cite{qamsane2019unified}. Latency-aware orchestration between edge and cloud halves response time with adaptive migration/load balancing in dynamic environments \cite{bulej2021managing}. In clinical operations, integrating ML and optimization has increased OR throughput and utilization by $15$--$25\%$ over manual planning \cite{ElSaddik2018DTMultimedia}. 

Across settings, modularity, parallelism, and GPU acceleration remain essential for real-time surgical planning, intraoperative feedback, and multi-user DT collaboration \cite{servin2023interactive},\cite{filippidis2024vr},\cite{servin2024simulation},\cite{servin2025digital},\cite{klimo2023digital},\cite{fuller2020digital},\cite{santoso2024maximizing}.

\section{Validation and Verification}

Robust validation and verification (V\&V) are indispensable to the clinical translation and reliability of digital twin (DT) technologies in healthcare. This section synthesizes the latest advances in validation methodologies, performance metrics, clinical validation, and ethical considerations, using concrete examples and quantitative data from high-impact studies.

\subsection{Methods of Validation in Digital Twins}
Validation in digital twin systems often starts with quantifying accuracy and reliability. In cardiac healthcare digital twins, Rudnicka et al.\ \cite{rudnicka2024cardiac} highlight approaches where predictive accuracy is benchmarked against established clinical data, with AI-based ECG models reaching up to $85.77\%$ accuracy in arrhythmia detection and risk stratification. Cockrell et al.\ \cite{cockrell2024weabm} report agent-based wound environment digital twins (WEABM) validated with $>95\%$ model fit to experimental healing curves, demonstrating high simulation fidelity. In the context of cardiac conduction modeling, Reza et al.\ \cite{reza2024assessing} evaluate digital twin predictions of post-TAVR conduction abnormalities using patient-specific simulations, achieving concordance with observed clinical outcomes in multiple prospective case studies. Tortora et al.\ \cite{tortora2025medical} note that model validation increasingly uses cross-validation with external datasets, with cardiac digital twins for sudden cardiac death outperforming traditional risk stratification tools in AUC and sensitivity.

Benchmarking techniques are employed extensively. Fairley et al.\ \cite{fairley2019improving} provide a strong example in surgical scheduling: their machine learning-based prediction of PACU recovery times is validated through discrete event simulation and compared to manual scheduling. Their approach led to a $76\%$ reduction in PACU holds and an $80\%$ reduction in PACU admission delays in a real-world hospital dataset (4{,}350 cases), while maintaining or improving operating room (OR) utilization. Benchmarking is also applied in resource management and ICU digital twins (Halpern et al.\ \cite{halpern2025advances}), where DT performance is evaluated by comparing clinical decision support recommendations with standard of care and alternative AI models, with some DTs showing improved time to sepsis detection and reduced ventilator-induced lung injury.

Kabir et al.\ \cite{kabir2025digital} provide a cross-domain summary of validation metrics in digital twins for smart healthcare, reporting device-level digital twins achieving $>96\%$ simulation fidelity and cardiovascular risk prediction models reaching $88.91\%$ accuracy on external cohorts. Data-driven benchmarking is often extended by employing multi-institutional or open datasets, as in large-scale cardiology cohorts or the multicenter ICU simulation trials reviewed by Halpern et al.\ \cite{halpern2025advances}. Sun et al.\ \cite{sun2023digital} emphasize the value of external benchmarking against real clinical endpoints to demonstrate generalizability and robustness.

\subsection{Performance Metrics for Emergency Decision-Making}
Performance evaluation in DT-assisted emergency decision-making requires a multidimensional approach. Key metrics include prediction accuracy, operational improvements, clinical impact, and workflow efficiency. Fairley et al.\ \cite{fairley2019improving} used a custom “operational accuracy” metric that penalizes deviations in predicted versus observed PACU durations, with a threshold mapping: 
\begin{align}
\tau(Y) &= \min\{\max\{\gamma Y, m\}, M\}\\
\gamma &= 0.2,\; m = 15\,\text{min},\; M = 60\,\text{min}.
\end{align}
They demonstrated a significant reduction in PACU congestion without compromising OR throughput.

Kabir et al.\ \cite{kabir2025digital} summarize DT-driven ER and hospital triage systems, reporting up to $80\%$ accuracy in real-time patient identification and resource allocation, with smart device digital twins enabling $20\%$ latency reduction and $>50\%$ power savings. In critical care, Halpern et al.\ \cite{halpern2025advances} found that DTs for sepsis and shock achieved higher early-warning sensitivity and specificity than traditional EHR alerts in simulated environments, although most DTs are at prototype or pilot stage, with further large-scale validation needed.

In stroke management, DT-based decision support tools have been evaluated using both accuracy and “time-to-intervention” metrics, showing that DT-driven care models can accelerate treatment compared to current clinical workflows \cite{halpern2025advances}. Kabir et al.\ \cite{kabir2025digital} further document that meta-learning digital twins for device performance monitoring reached $>96\%$ fidelity in hardware simulation environments, illustrating the high standards achievable in technical benchmarking.

\subsection{Case Studies and Experimental Results}
Real-world case studies and trials provide the strongest evidence for the impact of validated digital twin technologies. Fairley et al.\ \cite{fairley2019improving} present a six-month deployment at Stanford Children’s Hospital, where ML-optimized OR schedules improved operational efficiency by up to $25\%$ and reduced adverse bottlenecks. In cardiac applications, Reza et al.\ \cite{reza2024assessing} show that digital twin models can predict post-TAVR conduction abnormalities with high accuracy, validated through patient-specific simulations in clinical settings.

Rudnicka et al.\ \cite{rudnicka2024cardiac} describe prospective studies where cardiac digital twins are used for therapy response prediction, with cross-validation AUCs surpassing conventional predictors. Kabir et al.\ \cite{kabir2025digital} aggregate evidence from ER, ICU, and smart hospital deployments, documenting up to $80\%$ accuracy for DT-based ER triage and significant resource optimization gains. Halpern et al.\ \cite{halpern2025advances}, reviewing recent DT studies in critical care, highlight examples of improved sepsis detection, ventilator settings, and trainee education, while emphasizing the current gap in robust, multi-institutional clinical validation for most DTs.

Cockrell et al.\ \cite{cockrell2024weabm} document the WEABM wound-healing digital twin’s $>95\%$ match with in vivo time-course data, providing a rigorous example of simulation-based validation. In cancer care, Sun et al.\ \cite{sun2023digital} and Tortora et al.\ \cite{tortora2025medical} report digital twin-enabled personalization of oncology treatment, with model predictions of tumor growth validated against patient-specific imaging and response data.

\subsection{Ethical and Safety Validation}
Ethical and safety considerations are integral to digital twin validation, especially as AI and data-driven models proliferate. Kabir et al.\ \cite{kabir2025digital} and Tortora et al.\ \cite{tortora2025medical} both stress the importance of privacy, data security, and algorithmic transparency. Halpern et al.\ \cite{halpern2025advances} highlight the need for bias assessment, informed-consent protocols, and mechanisms for patient override or human-in-the-loop supervision, particularly in high-stakes critical care applications.

Regulatory standards are also emerging as a validation requirement. Halpern et al.\ \cite{halpern2025advances} reference the lag between rapid DT/AI development and regulatory adaptation (e.g., FDA pathways for software as a medical device). They and Tortora et al.\ \cite{tortora2025medical} recommend that multidisciplinary oversight—including ethicists, clinicians, and patients—be incorporated in the validation and deployment of digital twins, ensuring equity, safety, and trustworthiness.

Bias audits, as described in Kabir et al.\ \cite{kabir2025digital} and Rudnicka et al.\ \cite{rudnicka2024cardiac}, are increasingly required, with DT validation protocols now including fairness testing and adverse-event tracking. Sun et al.\ \cite{sun2023digital} suggest that DT frameworks should mandate periodic reviews of ethical and safety performance metrics, including transparency in data provenance and auditability of AI/ML decisions.

\section{Challenges and Open Issues}
\label{section10:Challenges and Open Issues}

The clinical translation of digital twin (DT) technology in healthcare is shaped by ongoing challenges across technical, clinical, ethical, and data domains. This section synthesizes recent findings and comparative data to clarify these persistent issues.

\subsection{Technical and Computational Challenges}
\label{sec:10.1}
Achieving real-time performance remains a core technical barrier for DT deployment. Laaki et al.\ reported that remote surgical DT control via 5G can achieve $<100\,\mathrm{ms}$ round-trip latency in best-case scenarios, yet even brief network jitter can jeopardize safety-critical feedback \cite{laaki2019prototyping}. Suraci et al.\ found that migrating DT computation to the edge reduces network latency by $20$--$30\%$ compared to cloud-only models, but adds orchestration complexity and potential data-consistency challenges \cite{suraci2025migrate}. Bulej et al.\ showed that edge--cloud environments can halve system response times under high loads via adaptive workload partitioning \cite{bulej2021managing}.

Integration and compatibility issues are significant. Chen et al.\ and Wang \& Luo highlight the need for open, standardized APIs and ontologies \cite{chen2024networking,wang2021digital}. In multi-vendor pilots, data harmonization and interoperability failures can account for substantial portions of system downtime \cite{chen2024networking,wang2021digital}. Hybrid architectures alleviate some challenges but require robust synchronization and redundancy mechanisms \cite{qamsane2019unified,santoso2024maximizing}.


\begin{table}[H]
\centering
\caption{Technical challenges across deployment styles.}
\label{tab:tech_challenges_10_1}
\renewcommand{\arraystretch}{1.2}
\begin{tabularx}{\columnwidth}{@{}p{0.1\textwidth} X X X@{}}
\toprule
\textbf{Aspect} & 
\makecell{\textbf{Cloud-only DT} \\ \cite{suraci2025migrate,bulej2021managing}} & 
\makecell{\textbf{Edge-enabled DT} \\ \cite{suraci2025migrate,bulej2021managing}} & 
\makecell{\textbf{Integrated IoT} \\ \cite{chen2024networking,wang2021digital}} \\
\midrule
Latency (ms) & $50$--$500$ & $1$--$50$ & Device-dependent; can be $<10$ \\
Data privacy & Higher risk (centralized) & Improved (more local processing) & Varies by standard/protocol \\
Interoperability & Moderate; vendor lock-in & Higher; more APIs required & Challenging; heterogeneous stacks \\
System downtime (typical cause) & Network bottlenecks & Edge coordination, device failures & Data/format mismatches, device diversity \\
Scalability & Excellent (centralized) & Good (needs orchestration) & Variable; often a major challenge \\
\bottomrule
\end{tabularx}
\end{table}

\subsection{Clinical Acceptance and Integration}
\label{sec:10.2}
Clinical uptake of surgical DTs is shaped by demonstrable benefit within existing workflows, onboarding burden, and institutional trust. In cardiovascular practice, organ-level DTs that combine anatomy, physiology, and device interaction have been incorporated by roughly $11\%$ of surveyed centers despite $>85\%$ perceived future value; facilitators include transparent assumptions and effective visualization during multidisciplinary reviews, while routine use hinges on PACS/EHR integration and standardized reporting \cite{rudnicka2024cardiac,corralacero2020digital,gong2023interactive}. 

In energy-based oncology interventions, biophysical twins supporting microwave ablation forecasting have progressed from phantoms and retrospective studies toward prospective evaluation; adoption is gated by patient-specific calibration time and coordinating intraoperative updates with imaging/guidance systems \cite{servin2023interactive,servin2024simulation,servin2025digital}. Training and rehearsal constitute a second pathway to acceptance: context-aware assistance and VR curricula report $20$--$40\%$ faster skill acquisition in pilots, with barriers in simulation-to-practice transfer, onboarding, and sustaining lab/OR support \cite{filippidis2024vr,shu2023twins,hagmann2021digital}. In perioperative operations, a reported $76\%$ reduction in PACU holds underscores measurable service-level impact, yet sites described shadow periods, role-specific training, and protocol updates for durable adoption \cite{fairley2019improving}. Telesurgery and network-aware twins show feasibility under intermittent communication by maintaining synchronized state with buffering/prediction, but governance typically requires evidence of fail-safes, handovers, and clear accountability before live deployment \cite{wang2025digital,bonne2022digital}. Reviews emphasize that prospective, multi-center evidence and post-deployment monitoring are central to institutional endorsement \cite{ringeval2025advancing,halpern2025advances}.

\begin{table}[htbp]
\centering
\caption{Clinical adoption vignettes and evidence points.}
\label{tab:clinical_adoption_10_2}
\renewcommand{\arraystretch}{1.2}
\begin{tabularx}{\columnwidth}{@{}p{0.08\textwidth} X X X@{}}
\toprule
\textbf{Setting} & \textbf{Deployment objective} & \textbf{Selected evidence} & \textbf{Principal integration considerations} \\
\midrule
Cardiovascular planning twins \cite{rudnicka2024cardiac,corralacero2020digital,gong2023interactive} & Patient-specific planning and MDT support & Adoption $\approx 11\%$; perceived value $>85\%$ & PACS/EHR integration; standardized reporting \\
Thermal ablation twins \cite{servin2023interactive,servin2024simulation,servin2025digital} & Forecast ablation volume/ thermal spread & Phantom/ retrospective $\rightarrow$ prospective transition & Calibration workload; intra-op update latency \\
Robotic training \& rehearsal \cite{filippidis2024vr,shu2023twins,hagmann2021digital} & Skills acquisition and rehearsal & Skill acquisition $-20$--$40\%$ & Simulation-to-practice transfer; onboarding/ support \\
Perioperative operations \cite{fairley2019improving} & Flow optimization and holds reduction & PACU holds $-76\%$ & Staff buy-in; KPI monitoring; protocol updates \\
Telesurgery/ network-aware DTs \cite{wang2025digital,bonne2022digital} & Safe guidance under variable QoS & Feasibility with buffering/prediction & Fail-safes; handovers; accountability \\
\bottomrule
\end{tabularx}
\end{table}

\subsection{Ethical, Legal, and Regulatory Considerations}
\label{sec:10.3}
DTs that influence diagnosis or treatment typically follow SaMD pathways (US) or MDR Rule~11 (EU) with risk-based classification, clinical evaluation, and lifecycle control tied to intended use \cite{ringeval2025advancing,sel2025survey,sendak2020path,morley2020ethics}. Human-factors engineering, alarm management, safe fallbacks, and operator-in-the-loop controls are emphasized for intraoperative/time-critical contexts \cite{hagmann2021digital,ding2024unifying,sendak2020path}. GDPR obligations apply whenever DTs process special-category health data; practical deployments respond with DPIAs, purpose limitation/minimization, and privacy-preserving designs such as federated learning and secure edge processing \cite{chen2024networking,ringeval2025advancing,kabir2025digital,leslie2019understanding}.

Evidence reviews highlight gaps relevant to ethics/compliance: a sizable fraction of DT pilots lack end-to-end encryption despite handling sensitive multi-source clinical data; recommended mitigations include encryption-by-default, hardened IAM, and edge segmentation \cite{kabir2025digital,halpern2025advances}. Fairness and explainability remain concerns; in cardiac DT audits, up to $18\%$ of adverse recommendations were linked to unbalanced training data, motivating bias assessment, dataset curation, and clinically aligned explanation modalities \cite{rudnicka2024cardiac}. Regulatory analyses call for post-market surveillance with real-world performance tracking, drift monitoring, and corrective actions proportionate to clinical risk, particularly for adaptive or locally retrained components \cite{ringeval2025advancing,halpern2025advances,morley2020ethics}.

\begin{table*}[htbp]
\centering
\caption{Regulatory \& governance checklist for surgical digital twins.}
\label{tab:regulatory_checklist_10_3}
\renewcommand{\arraystretch}{1.2}
\begin{tabularx}{\textwidth}{@{}p{0.26\textwidth} X X@{}}
\toprule
\textbf{Dimension} & \textbf{Oversight expectations} & \textbf{Practical artifacts for SDTs} \\
\midrule
Intended use \& risk \cite{ringeval2025advancing,sendak2020path,morley2020ethics} & Claims aligned to SaMD/MDR class \& context & Intended-use statement; hazard analysis; workflow mapping \\
Clinical evaluation \cite{ringeval2025advancing,sel2025survey,halpern2025advances} & Evidence that DT outputs improve decisions/outcomes & Protocols, endpoints, prospective/multi-site studies \\
VVUQ \& lifecycle control \cite{ringeval2025advancing,sel2025survey} & Verification/validation; uncertainty \& change control & Model cards; drift monitors; version/change logs \\
Human factors \& safety \cite{hagmann2021digital,ding2024unifying,sendak2020path} & Operator-in-loop, alarms, safe fallbacks, usability & HFE reports; summative testing; safety case \\
Interoperability \cite{chen2024networking,qamsane2019unified} & Robust interfaces and error handling & API specs; ontology/data mappings; interface tests \\
Cybersecurity \cite{chen2024networking,kabir2025digital,leslie2019understanding} & Confidentiality, integrity, availability (edge/cloud) & Threat models; end-to-end encryption; incident response \\
Data protection (GDPR) \cite{chen2024networking,ringeval2025advancing,kabir2025digital,leslie2019understanding} & DPIA; minimization; transfer safeguards & DPIA; retention policy; federated/edge designs \\
Post-market monitoring \cite{ringeval2025advancing,halpern2025advances,morley2020ethics} & Real-world performance, bias, safety oversight & RWE plans; fairness audits; CAPA records \\
\bottomrule
\end{tabularx}
\end{table*}

\subsection{Data Quality and Availability}
\label{sec:10.4}
The quality, diversity, and annotation of data underpin every phase of DT development. Data heterogeneity—formats, devices, and EHR systems—remains a top challenge \cite{mihai2022digital,wang2021digital,alauddin2018bibliometric,bhatia2016temporal}. Only about $23\%$ of surveyed health systems in Kabir et al.\ reported standardized DT-ready data pipelines \cite{kabir2025digital}. Scarcity is acute in rare-disease modeling: Sun et al.\ note that $<5\%$ of pediatric centers maintain sufficiently large, labeled datasets for AI-driven DT research \cite{sun2023digital}. 

Annotation quality and consistency are limiting factors: Blasch et al.\ found that up to $32\%$ of ML errors in sensor fusion were linked to mislabelled or missing data \cite{blasch2021ml}. Multi-institutional data-sharing initiatives (e.g., large cardiology cohorts; open ICU DT benchmarks) are critical steps but require sustained collaboration and standard ontologies \cite{sun2023digital,rudnicka2024cardiac,halpern2025advances}.

\begin{table}[http]
\centering
\caption{Methods to address data quality and availability in healthcare DTs.}
\label{tab:data_quality_methods_10_4}
\renewcommand{\arraystretch}{1.2}
\begin{tabularx}{\columnwidth}{@{}p{0.125\textwidth} p{0.1\textwidth} X X@{}}
\toprule
\textbf{Method/ approach} & \textbf{Core idea/ example} & \textbf{Reported benefit/ statistic} & \textbf{Limitation/ challenge} \\
\midrule
Data harmonization \& standardization \cite{mihai2022digital,chen2024networking,wang2021digital,kabir2025digital} & Common data models \& ontologies & Reduces downtime from data mismatches & Slow adoption; consensus required \\
Federated learning \cite{blasch2021ml,kabir2025digital} & Collaborative training without raw data sharing & Preserves privacy; increases accessible data & Requires infrastructure; systems integration \\
Multi-institutional data sharing \cite{sun2023digital,rudnicka2024cardiac,halpern2025advances} & Large cross-site cohorts; shared benchmarks & Improves sample size \& diversity & Regulatory/legal barriers \\
Annotation quality control \cite{blasch2021ml} & Auditing; consensus annotation & Up to $32\%$ fewer ML errors in fusion & Labor-intensive; subjectivity risk \\
\bottomrule
\end{tabularx}
\end{table}

\section{Future Research Directions}

The evolving landscape of digital twin (DT) technology in healthcare—especially in surgical and precision medicine applications—presents a vibrant field for ongoing research. Building on advances in communication, modeling, validation, and clinical implementation, recent studies highlight next-generation technologies, broader clinical integration, interdisciplinary collaboration, and economic considerations.

\subsection{Emerging Technologies and Innovations}
\label{sec:11.1}
The convergence of artificial intelligence, advanced sensing, and high-speed networks is poised to drive the next wave of DT innovations. Rajkomar et al.\ note that machine learning and deep learning continue to increase prediction accuracy and automation across diagnostic and prognostic tasks \cite{rajkomar2019machine}, while Kamel Boulos \& Zhang emphasize the growing role of IoT and big-data analytics in both individual and population-level twins \cite{kamel2021digital}. Reviews such as Ayd{\i}n \& Karaarslan catalog rapid progress integrating AI/ML with DTs to enhance predictive and adaptive capabilities \cite{aydin2022openai}. Coorey et al.\ showcase cardiovascular DTs leveraging continuous EHR and wearable streams for real-time monitoring and dynamic therapy adjustment \cite{coorey2022health}.

Emerging technologies include 6G communication, edge AI, and blockchain for real-time, secure, privacy-preserving DT ecosystems \cite{chen2024networking,kamel2021digital,aydin2022openai}. For ultra-low-latency medical applications, edge-enabled DTs and future 6G systems are promising \cite{chen2024networking,suraci2025migrate}. Integration of multimodal sensor data is enabling holistic twins that not only predict acute events but also adapt with new patient data over time \cite{rajkomar2019machine,coorey2022health}.

\begin{table}[http]
    \centering
    \caption{Key technologies driving the next generation of digital twins in healthcare.}
    \label{tab:sec11_keytech_new}
    \renewcommand{\arraystretch}{1.15}
    \begin{tabularx}{\columnwidth}{@{}p{0.1\textwidth} p{0.14\textwidth} p{0.21\textwidth}@{}}
    \toprule
    \textbf{Technology} & \textbf{Main advantage} & \textbf{Example use cases / findings} \\
    \midrule
    AI \& Deep Learning \cite{rajkomar2019machine,rudnicka2024cardiac,coorey2022health} & High prediction accuracy; adaptive models & Diagnosis \cite{rajkomar2019machine}; risk stratification \cite{rudnicka2024cardiac}; outcome prediction \cite{coorey2022health} (e.g., arrhythmia detection $\approx 85.77\%$) \\
    IoT Integration \cite{kamel2021digital,aydin2022openai,coorey2022health} & Real-time, continuous, multi-modal data & Remote monitoring \cite{kamel2021digital}; precision public health \cite{aydin2022openai}; wearables integration \cite{coorey2022health} \\
    Edge Computing \cite{suraci2025migrate,chen2024networking,kamel2021digital} & Ultra-low latency; privacy preservation & Near-device computation \cite{suraci2025migrate}; scalable DT operations \cite{chen2024networking}; $<1$\,ms targets in 6G roadmaps \cite{kamel2021digital} \\
    Blockchain \cite{kamel2021digital,aydin2022openai} & Data security; traceability & Privacy-preserving data sharing across institutions \\
    \bottomrule
    \end{tabularx}
\end{table}

\subsection{Expanding Digital Twin Applications in Surgery}
\label{sec:11.2}
As DT models mature, their scope extends beyond single-procedure rehearsal to broader perioperative and cross-disciplinary care. VR/AR-based DT platforms support surgical skill training, anatomy simulation, and intraoperative guidance, improving procedural accuracy and learning outcomes \cite{shu2023twins,klimo2023digital}. Pilot studies in personalized planning report reduced planning time and improved outcomes \cite{cellina2023digital}. Scaling from individualized surgery to precision public health is enabled by combining genomics, imaging, and sensors \cite{kamel2021digital}; cardiology DTs integrate imaging, genetics, and real-time clinical data \cite{coorey2022health}, while ethical/technical complexities call for interdisciplinary engagement \cite{bruynseels2018digital}.

Expanding use-cases include real-time OR scheduling \cite{fairley2019improving}, critical care \cite{halpern2025advances}, smart device monitoring \cite{kabir2025digital}, and personalization in oncology and chronic disease management \cite{sun2023digital,tortora2025medical,coorey2022health}. Success hinges on interoperability standards, robust validation, and collaboration across medical, computational, and engineering domains \cite{chen2024networking,aloqaily2023digital,bruynseels2018digital,coorey2022health}.

\subsection{Interdisciplinary Research Opportunities}
\label{sec:11.3}
Progress will hinge on co-design among clinicians, data scientists, ethicists, and engineers \cite{bruynseels2018digital,kamel2021digital}. Academia–industry partnerships can accelerate translation, provided ethical and regulatory frameworks keep pace \cite{rajkomar2019machine,aydin2022openai}. Privacy-preserving analytics (e.g., federated learning) enable multi-center studies without raw data sharing \cite{chen2024networking,bruynseels2018digital,aydin2022openai}. Multicenter clinical studies (e.g., cardiology cohorts) are crucial to validate DTs and drive standardization of architectures and data models \cite{coorey2022health}.
\begin{figure}
    \centering
    \includegraphics[width=\linewidth]{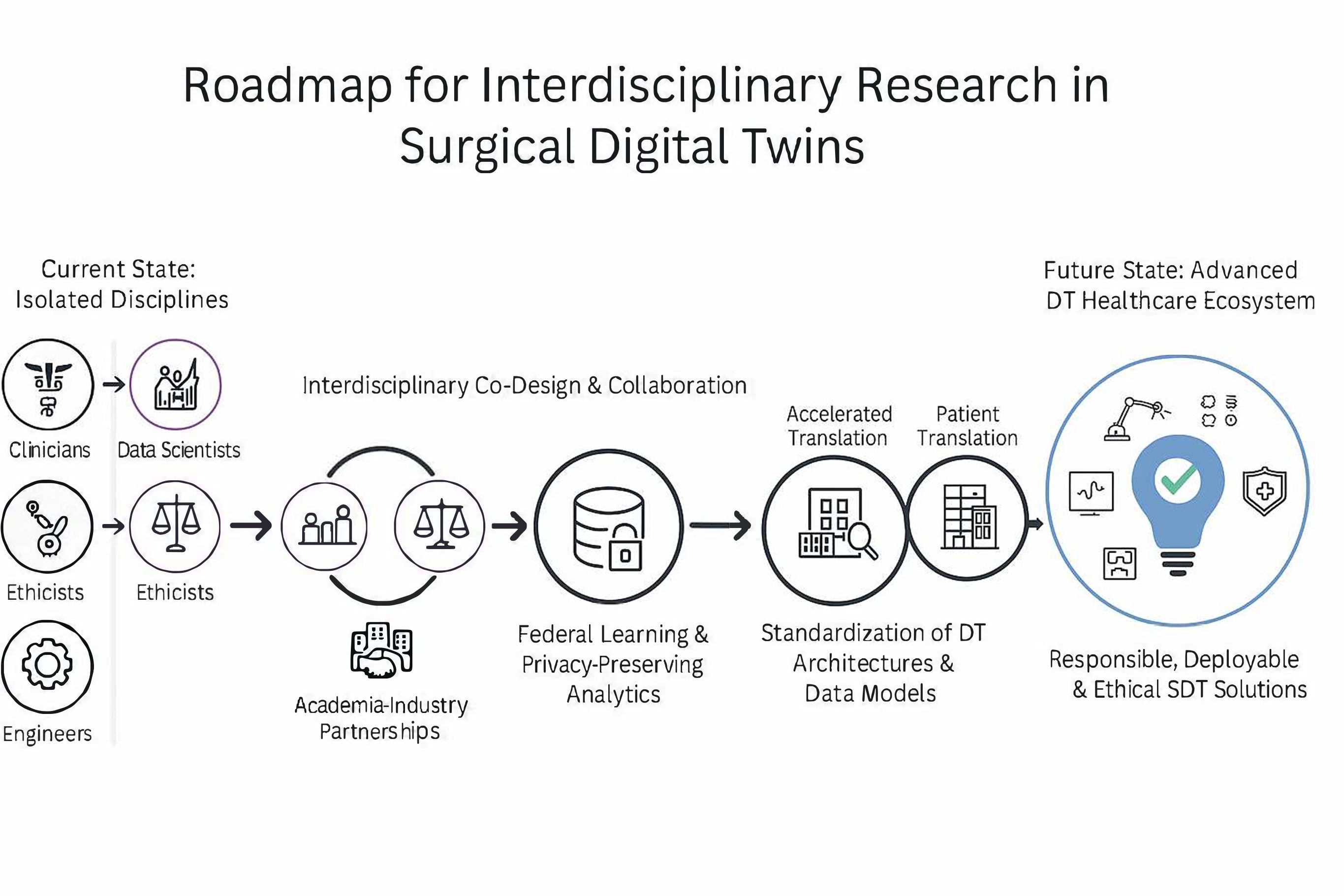}
    \caption{A Roadmap for Interdisciplinary Collaboration in Surgical Digital Twin (SDT) Research.}
    \label{fig:placeholder}
\end{figure}

\subsection{Economic and Operational Impact}
\label{sec:11.4}
Quantifying economic and operational benefits is an emerging focus. Early deployments report improved efficiency (reduced planning/intervention time), patient safety, and cost savings via better resource utilization and prevention of adverse events \cite{cellina2023digital,coorey2022health}. Smart device twins show reduced triage latency and power consumption \cite{kabir2025digital}. In perioperative operations, DT-based scheduling achieved $76\%$ fewer PACU holds and up to $25\%$ higher OR throughput \cite{fairley2019improving}. Future studies should rigorously assess ROI and cost-effectiveness, as up-front DT investments may be offset by long-term savings and quality gains \cite{rajkomar2019machine,kamel2021digital}.

\begin{table}[htbp]
\centering
\caption{Selected economic/operational outcomes from DT adoption.}
\label{tab:sec11_econ_new}
\renewcommand{\arraystretch}{1.15}
\begin{tabularx}{\columnwidth}{@{}p{0.4\columnwidth} X@{}}
\toprule
\textbf{Study / setting} & \textbf{Measured benefit} \\
\midrule
Hospital OR (Stanford) \cite{fairley2019improving} & $76\%$ fewer PACU holds; $80\%$ fewer admission delays; up to $25\%$ higher throughput \\
Device digital twins \cite{kabir2025digital} & Up to $20\%$ latency reduction; $50\%+$ power savings \\
Cardiology DTs \cite{coorey2022health,cellina2023digital} & Shorter planning times; improved outcomes \\
General cost--benefit \cite{rajkomar2019machine,kamel2021digital} & ROI possible with long-term scaling (ongoing study) \\
\bottomrule
\end{tabularx}
\end{table}

\begin{table*}[htbp]
\centering
\caption{Challenge--future solution crosswalk linking Sections~10 and~11.}
\label{tab:sec11_crosswalk_new}
\renewcommand{\arraystretch}{1.15}
\begin{tabularx}{\textwidth}{@{}p{0.3\textwidth} p{0.3\columnwidth} X@{}}
\toprule
\textbf{Challenge in Sec. \ref{section10:Challenges and Open Issues}} & \textbf{References} & \textbf{Suggested approach / opportunity} \\
\midrule
Real-time constraints under variable QoS and heavy simulation loads (\ref{sec:10.1}) & \ref{sec:11.1} \cite{wang2025digital,suraci2025migrate,chen2024networking,bonne2022digital,bulej2021managing} & Edge-first partitioning with explicit latency budgets; QoS-aware scheduling; network-aware synchronization for telesurgery; resilience to intermittency \\
Interoperability and vendor heterogeneity across imaging, sensors, robots, IT (\ref{sec:10.1}) & \ref{sec:11.1} / \ref{sec:11.2} \cite{mihai2022digital,chen2024networking,qi2018digital,fuller2020digital,qamsane2019unified,lu2020digital} & Common data models/ontologies; open interface specs; reference DT stacks; conformance/interoperability test suites \\
Clinical acceptance and workflow integration (\ref{sec:10.2}) & \ref{sec:11.2} / \ref{sec:11.3} \cite{ringeval2025advancing,halpern2025advances,hagmann2021digital,fairley2019improving,rajkomar2019machine} & Prospective multi-site evaluations; human-factors evidence with operator-in-the-loop; ROI/KPI tracking for service adoption \\
Ethical, legal, regulatory alignment (\ref{sec:10.3}) & \ref{sec:11.3} \cite{ringeval2025advancing,sel2025survey,morley2020ethics,chen2024networking,kabir2025digital,leslie2019understanding} & SaMD/MDR-conformant VVUQ; post-market monitoring/drift control; DPIA-driven privacy; secure edge deployments \\
Data quality, heterogeneity, annotation burden (\ref{sec:10.4}) & \ref{sec:11.1} / \ref{sec:11.3} \cite{mihai2022digital,chen2024networking,rudnicka2024cardiac,blasch2021ml,sun2023digital,servin2024simulation,servin2025digital,halpern2025advances} & Standardized pipelines/harmonization; federated/split learning; multi-institution sharing; annotation QA; hybrid physics--ML surrogates with uncertainty reporting \\
\bottomrule
\end{tabularx}
\end{table*}

\section{Conclusions}

Digital twin technology has rapidly emerged as a transformative paradigm in healthcare, offering the promise of individualized, real-time, and predictive care by linking heterogeneous clinical, biological, and behavioral data streams into dynamically updated, patient-specific models. As this review demonstrates, the breadth of DT applications now spans precision medicine, personalized surgery, medical device and drug development, hospital operations, and longitudinal patient monitoring. Central to these advances is the capacity of DTs to facilitate seamless, bidirectional integration between the physical and virtual worlds, thereby enabling adaptive simulation, AI-driven intervention, and data-driven decision support within increasingly complex clinical workflows. 

Despite this progress, the field is still marked by persistent challenges. Issues related to the acquisition and harmonization of diverse data sources, the validation and regulatory acceptance of computational models, system interoperability, and concerns about privacy, equity, and governance remain formidable. Successful translation of digital twins from experimental to routine clinical use will require not only continued technological innovation, but also the establishment of robust frameworks for data quality, security, and societal trust.

When viewed in the context of existing literature, this review builds on and extends the work of several recent General \& Surgical Digital Twin Surveys. Notably, Ding et al.\ advanced the discussion by proposing a modular and interpretable DT framework for surgical data science, focusing on geometric scene understanding and algorithmic limitations \cite{ding2024unifying}; however, their analysis is confined primarily to computer vision and surgical informatics. The survey by Sun et al.\ provides a broad yet somewhat selective overview of DT applications across medicine, with limited integration of regulatory, societal, and technical dimensions \cite{sun2023digital}. In a distinct vein, Bruynseels et al.\ undertake an ethical and conceptual analysis, exploring how digital twins may redefine central medical concepts such as normality, diagnosis, and enhancement, but without mapping technical architectures or real-world barriers to implementation \cite{bruynseels2018digital}. Additional reviews by Asciak et al., Katsoulakis et al., and Abd~Elaziz et al.\ contribute further cataloging of DT types, application domains, and emerging research trends, though they often emphasize technical or simulation aspects while providing less coverage of governance, standards, and cross-domain coordination \cite{asciak2025digital,katsoulakis2024digital,abdelaziz2024digital}.

In parallel with these broad surveys, there has been a noticeable emergence of Specialty-Specific Surgical Digital Twin Reviews that probe the use of DT methods within distinct clinical domains. Seth et al.\ present a systematic review of DT applications in plastic surgery, highlighting advancements in patient-specific simulation, surgical planning, and improved clinical outcomes, but with less attention to broader system-level, regulatory, or societal challenges \cite{seth2024digital}. Bjelland et al.\ deliver a comprehensive analysis of DT-enabling technologies for arthroscopic knee surgery, evaluating progress in biomechanical modeling, haptic feedback, and intraoperative guidance, yet similarly constrained to a focused application area \cite{bjelland2022toward}. Corral-Acero et al.\ provide an authoritative position on DTs for precision cardiology, emphasizing the synergy of mechanistic and statistical modeling for individualized patient care, but concentrating on cardiovascular systems to the exclusion of broader healthcare integration \cite{corralacero2020digital}.

While these prior surveys each contribute important domain-specific or methodological insights, there is a clear need for a unified, multidisciplinary synthesis that systematically examines the technical architectures, modeling approaches, clinical workflows, regulatory landscapes, and societal implications of digital twins in health. The present review distinguishes itself through its breadth and depth, bringing together technical, clinical, ethical, and policy perspectives in a cohesive analysis. Unlike earlier works that tend to focus narrowly on either engineering or ethical issues—or which are restricted to particular specialties—this review addresses the full ecosystem of digital twins for health, from static and mirror to shadow and intelligent types, and maps them onto diverse healthcare scenarios such as hospital command centers, remote patient management, and \textit{in silico} clinical trials.

Importantly, this synthesis also foregrounds the role of global research consortia, industry initiatives, and emerging standards—exemplified by the Swedish Digital Twin Center, DIGIPREDICT, PRIMAGE, and the Digital Twin Consortium—in shaping the development and deployment of digital twin technology. By integrating examples and lessons from international collaboration, this review underscores the necessity of harmonized development strategies, robust governance frameworks, and patient-centered design principles for responsible and impactful innovation.

Ultimately, the adoption of digital twins in healthcare presents profound opportunities as well as new forms of risk. On one hand, DTs offer pathways to democratize high-quality care, accelerate biomedical research, and optimize system-level delivery. On the other, they introduce significant challenges related to privacy, digital inclusion, regulatory oversight, and the distribution of technological benefit. The realization of the transformative promise of digital twins will depend not only on sustained technical progress, but also on transparent governance, public engagement, and the fostering of trust across all stakeholders.

In summary, this review advances the state of the art by offering a comprehensive, interdisciplinary, and forward-looking assessment of digital twin technology in healthcare. Through its integrated treatment of technical, clinical, ethical, and societal dimensions, it provides a roadmap for future research and clinical translation, highlighting the continued importance of inclusivity, interoperability, and robust standards as foundational elements for the next generation of digital health innovation.

\appendices
\appendix
\section{DEVS Formal Definitions \& Notation}

\subsection{Atomic model (classic DEVS)}
An atomic model is
\begin{align}
M &= \langle X, S, Y, \delta_{\mathrm{int}}, \delta_{\mathrm{ext}}, \lambda, t_a \rangle, \\
\delta_{\mathrm{int}} &:\; S \to S, \\
\delta_{\mathrm{ext}} &:\; S \times X \times \mathbb{R}_{\ge 0} \to S, \\
\lambda &:\; S \to Y, \\
t_a &:\; S \to \mathbb{R}_{\ge 0}.
\end{align}
Execution alternates between waiting $t_a(s)$, emitting $\lambda(s)$, and applying $\delta_{\mathrm{int}}$; inputs arriving before $t_a$ trigger $\delta_{\mathrm{ext}}$ \cite{zeigler2000theory}.

\subsection{Coupled model \& hierarchy}
A coupled model organizes atomic/coupled submodels with input/output couplings and \emph{select} functions to handle simultaneous events. Hierarchy is achieved by nesting coupled models; the closure-under-coupling property ensures any coupled model behaves like an atomic one, enabling scalable composition \cite{zeigler2000theory}.

\subsection{Co-simulation with continuous physics}
Between events, continuous solvers (e.g., FEA/bioheat) advance biophysical states; at events, interface variables exchange values with DEVS components (e.g., probe insertion, imaging update). Synchronization policies (discrete-event timestamps, lookahead) maintain causality under edge/cloud deployment and variable latency \cite{gong2023interactive,servin2023interactive,servin2024simulation,suraci2025migrate,zeigler2000theory}.

\subsection{Practical mapping to SDTs}
\begin{itemize}
  \item \textbf{Workflow / alarms:} map tasks, decisions, and alerts to atomic models; use couplings for team/tool communication \cite{asciak2025digital,hagmann2021digital}.
  \item \textbf{Data fusion:} treat sensor/vision streams as input events; gate continuous estimates with DEVS scheduling \cite{elayan2021digital,das2022toward}.
  \item \textbf{Telesurgery / networking:} represent link changes as events; elasticity policies (edge vs.\ cloud) are modeled as structural couplings \cite{xie2024tiodt,suraci2025migrate}.
\end{itemize}

\bibliographystyle{IEEEtran}   
\bibliography{reference}       

\begin{thebibliography}{100}
\providecommand{\url}[1]{#1}
\csname url@samestyle\endcsname
\providecommand{\newblock}{\relax}
\providecommand{\bibinfo}[2]{#2}
\providecommand{\BIBentrySTDinterwordspacing}{\spaceskip=0pt\relax}
\providecommand{\BIBentryALTinterwordstretchfactor}{4}
\providecommand{\BIBentryALTinterwordspacing}{\spaceskip=\fontdimen2\font plus
\BIBentryALTinterwordstretchfactor\fontdimen3\font minus \fontdimen4\font\relax}
\providecommand{\BIBforeignlanguage}[2]{{%
\expandafter\ifx\csname l@#1\endcsname\relax
\typeout{** WARNING: IEEEtran.bst: No hyphenation pattern has been}%
\typeout{** loaded for the language `#1'. Using the pattern for}%
\typeout{** the default language instead.}%
\else
\language=\csname l@#1\endcsname
\fi
#2}}
\providecommand{\BIBdecl}{\relax}
\BIBdecl

\bibitem{laaki2019prototyping}
H.~Laaki, Y.~Miche, and K.~Tammi, ``Prototyping a digital twin for real-time remote control over mobile networks: Application of remote surgery,'' \emph{IEEE Access}, 2019.

\bibitem{sun2023digital}
T.~Sun, X.~He, and Z.~Li, ``Digital twin in healthcare: Recent updates and challenges,'' \emph{Digital Health}, 2023.

\bibitem{elayan2021digital}
H.~Elayan, M.~Aloqaily, and M.~Guizani, ``Digital twin for intelligent context-aware iot healthcare systems,'' \emph{IEEE Internet of Things Journal}, 2021.

\bibitem{lonsdale2022perioperative}
H.~Lonsdale, G.~M. Gray, L.~M. Ahumada, H.~M. Yates, A.~Varughese, and M.~A. Rehman, ``The perioperative human digital twin,'' \emph{Anesthesia \& Analgesia}, vol. 134, no.~4, pp. 885--892, 2022.

\bibitem{servin2024simulation}
F.~Servin, J.~A. Collins, J.~S. Heiselman \emph{et~al.}, ``Simulation of image-guided microwave ablation therapy using a digital twin computational model,'' \emph{IEEE Open Journal of Engineering in Medicine and Biology}, 2024.

\bibitem{gong2023interactive}
Y.~Gong, F.~Qi, J.-Y. Wang \emph{et~al.}, ``An interactive platform for a high performance digital twin of a human heart,'' in \emph{2023 IEEE International Conference on Metaverse Computing, Networking and Applications (MetaCom)}, 2023.

\bibitem{annamraju2024digital}
S.~Annamraju, P.~Jeziorczak, Q.~Ye, and I.~Kim, ``Digitaltwin for surgical tool--tissue interaction: A systems perspective,'' in \emph{2024 Annual Modeling and Simulation Conference (ANNSIM)}, 2024.

\bibitem{tai2022digital}
Y.~Tai, J.~Hu, K.~Zhang, H.~Liu, and H.~Zhu, ``Digital-twin-enabled iomt system for surgical simulation using rac-gan,'' \emph{IEEE Internet of Things Journal}, 2022.

\bibitem{servin2025digital}
F.~Servin, J.~A. Collins, J.~S. Heiselman \emph{et~al.}, ``Digital twin modeling and machine learning frameworks for forecasting multiple microwave ablation volumes,'' in \emph{Medical Imaging 2025: Image-Guided Procedures, Robotic Interventions, and Modeling (SPIE)}, 2025.

\bibitem{chaudhuri2023predictive}
A.~Chaudhuri, G.~Pash, D.~A. Hormuth, G.~Lorenzo, M.~Kapteyn, C.~Wu, E.~A. Lima, T.~E. Yankeelov, and K.~Willcox, ``Predictive digital twin for optimizing patient-specific radiotherapy regimens under uncertainty in high-grade gliomas,'' \emph{Frontiers in Artificial Intelligence}, vol.~6, p. 1222612, 2023.

\bibitem{asciak2025digital}
L.~Asciak, J.~Kyeremeh, X.~Luo \emph{et~al.}, ``Digital twin assisted surgery, concept, opportunities, and challenges,'' \emph{npj Digital Medicine}, 2025.

\bibitem{shu2023twins}
H.~Shu, R.~Liang, Z.~Li, H.~Zhang, Y.~Zhang, M.~Xu, and F.~Meng, ``Twin-s: A digital twin for skull base surgery,'' \emph{International Journal of Computer Assisted Radiology and Surgery}, 2023.

\bibitem{das2022toward}
C.~Das, A.~A. Mumu, M.~F. Ali \emph{et~al.}, ``Toward iort collaborative digital twin technology enabled future surgical sector: Technical innovations, opportunities and challenges,'' \emph{IEEE Access}, 2022.

\bibitem{zhang2022artificial}
Z.~Zhang, F.~Wen, Z.~Sun, X.~Guo, T.~He, and C.~Lee, ``Artificial intelligence-enabled sensing technologies in the 5g/internet of things era: From vr/ar to the digital twin,'' \emph{Advanced Intelligent Systems}, 2022.

\bibitem{lippert2024cardiac}
M.~Lippert, K.-A. Dumont, S.~Birkeland, V.~Nainamalai, H.~Solvin, K.~R. Suther, B.~Bendz, O.~J. Elle, and H.~Brun, ``Cardiac anatomic digital twins: findings from a single national centre,'' \emph{European Heart Journal-Digital Health}, vol.~5, no.~6, pp. 725--734, 2024.

\bibitem{kleinbeck2024neural}
C.~Kleinbeck, H.~Zhang, B.~D. Killeen, D.~Roth, and M.~Unberath, ``Neural digital twins: reconstructing complex medical environments for spatial planning in virtual reality,'' \emph{International Journal of Computer Assisted Radiology and Surgery}, vol.~19, no.~7, pp. 1301--1312, 2024.

\bibitem{xie2024tiodt}
D.~Xie, J.~Diao, F.~Fang \emph{et~al.}, ``Tiodt: Touchfree intuitive operation digital twin platform for minimally invasive surgical robot,'' in \emph{2024 IEEE International Conference on Real-Time Computing and Robotics (RCAR)}, 2024.

\bibitem{bjelland2022toward}
{\O}.~Bjelland, B.~Rasheed, H.~G. Schaathun \emph{et~al.}, ``Toward a digital twin for arthroscopic knee surgery: A systematic review,'' \emph{IEEE Access}, 2022.

\bibitem{hagmann2021digital}
K.~Hagmann, A.~Hellings-Ku{\ss}, J.~Klodmann \emph{et~al.}, ``A digital twin approach for contextual assistance for surgeons during surgical robotics training,'' \emph{Frontiers in Robotics and AI}, 2021.

\bibitem{wang2025digital}
J.~Wang, J.~A. Barragan, H.~Ishida \emph{et~al.}, ``A digital twin for telesurgery under intermittent communication,'' in \emph{2025 International Symposium on Medical Robotics (ISMR)}, 2025.

\bibitem{perez2025privacy}
A.~Perez, H.~Zhang, Y.-C. Ku, L.~Seenivasan, R.~Soberanis, J.~L. Porras, R.~Day, J.~Jopling, P.~Najjar, and M.~Unberath, ``Privacy-preserving operating room workflow analysis using digital twins,'' \emph{arXiv preprint arXiv:2504.12552}, 2025.

\bibitem{cai2023implementation}
X.~Cai, Z.~Wang, S.~Li, J.~Pan, C.~Li, and Y.~Tai, ``Implementation of a virtual reality based digital-twin robotic minimally invasive surgery simulator,'' \emph{Bioengineering}, vol.~10, no.~11, p. 1302, 2023.

\bibitem{tarng2024application}
W.~Tarng, Y.-J. Wu, L.-Y. Ye, C.-W. Tang, Y.-C. Lu, T.-L. Wang, and C.-L. Li, ``Application of virtual reality in developing the digital twin for an integrated robot learning system,'' \emph{Electronics}, vol.~13, no.~14, p. 2848, 2024.

\bibitem{kaliappan2024digital}
D.~S. Kaliappan and S.~Logesh, ``Digital twin-driven augmented vision system for healthcare,'' in \emph{2024 9th International Conference on Communication and Electronics Systems (ICCES)}, 2024.

\bibitem{servin2023interactive}
F.~Servin, J.~A. Collins, J.~S. Heiselman \emph{et~al.}, ``Digital twin forecasting of microwave ablation via fat quantification image-to-grid computational methods,'' in \emph{Medical Imaging: Image-Guided Procedures, Robotic Interventions, and Modeling (SPIE)}, 2023.

\bibitem{chen2024networking}
J.~Chen, C.~Yi, S.~D. Okegbile, J.~Cai, and X.~Shen, ``Networking architecture and key supporting technologies for human digital twin in personalized healthcare: A comprehensive survey,'' \emph{IEEE Communications Surveys \& Tutorials}, 2024.

\bibitem{albertini2024digital}
J.-N. Albertini, L.~Derycke, A.~Millon, and R.~Soler, ``Digital twin and artificial intelligence technologies for predictive planning of endovascular procedures,'' in \emph{Seminars in Vascular Surgery}, vol.~37, no.~3.\hskip 1em plus 0.5em minus 0.4em\relax Elsevier, 2024, pp. 306--313.

\bibitem{filippidis2024vr}
A.~Filippidis, N.~Marmaras, M.~Maravgakis \emph{et~al.}, ``Vr isle academy: A vr digital twin approach for robotic surgical skill development,'' \emph{arXiv preprint arXiv:2406.00002}, 2024.

\bibitem{ding2024towards_automation}
H.~Ding, L.~Seenivasan, H.~Shu \emph{et~al.}, ``Towards robust automation of surgical systems via digital twin-based scene representations from foundation models,'' \emph{arXiv preprint arXiv:2409.13107}, 2024.

\bibitem{martinez2021machine}
O.~Martinez, C.~Martinez, C.~A. Parra, S.~Rugeles, and D.~R. Suarez, ``Machine learning for surgical time prediction,'' \emph{Computer Methods and Programs in Biomedicine}, 2021.

\bibitem{saxby2023digital}
D.~J. Saxby, C.~Pizzolato, and L.~E. Diamond, ``A digital twin framework for precision neuromusculoskeletal health care: extension upon industrial standards,'' \emph{Journal of applied biomechanics}, vol.~39, no.~5, pp. 347--354, 2023.

\bibitem{ding2024towards_scene}
H.~Ding, Y.~Zhang, W.~Cheng \emph{et~al.}, ``Towards robust algorithms for surgical phase recognition via digital twin--based scene representation,'' \emph{arXiv preprint arXiv:2410.20026}, 2024.

\bibitem{corralacero2020digital}
J.~Corral-Acero, F.~Margara, M.~Marciniak \emph{et~al.}, ``The ‘digital twin’ to enable the vision of precision cardiology,'' \emph{European Heart Journal}, 2020.

\bibitem{qin2022realizing}
J.~Qin and J.~Wu, ``Realizing the potential of computer-assisted surgery by embedding digital twin technology,'' \emph{JMIR Medical Informatics}, 2022.

\bibitem{katsoulakis2024digital}
E.~Katsoulakis, Q.~Wang, H.~Wu \emph{et~al.}, ``Digital twins for health: a scoping review,'' \emph{npj Digital Medicine}, 2024.

\bibitem{abdelaziz2024digital}
M.~Abd~Elaziz, M.~A.~A. Al-qaness, A.~Dahou \emph{et~al.}, ``Digital twins in healthcare: Applications, technologies, simulations, and future trends,'' \emph{WIREs Data Mining and Knowledge Discovery}, 2024.

\bibitem{ding2024unifying}
H.~Ding, L.~Seenivasan, B.~D. Killeen \emph{et~al.}, ``Digital twins as a unifying framework for surgical data science: the enabling role of geometric scene understanding,'' \emph{Artificial Intelligence Surgery}, 2024.

\bibitem{bruynseels2018digital}
K.~Bruynseels, F.~Santoni~de Sio, and J.~van~den Hoven, ``Digital twins in health care: Ethical implications of an emerging engineering paradigm,'' \emph{Frontiers in Genetics}, 2018.

\bibitem{blasch2021ml}
E.~Blasch, T.~Pham, C.-Y. Chong \emph{et~al.}, ``Machine learning/artificial intelligence for sensor data fusion--opportunities and challenges,'' \emph{IEEE Aerospace \& Electronic Systems Magazine}, 2021.

\bibitem{zeigler2000theory}
B.~P. Zeigler, H.~Praehofer, and T.~G. Kim, \emph{Theory of Modeling and Simulation}.\hskip 1em plus 0.5em minus 0.4em\relax Academic Press, 2000.

\bibitem{kuruppu2025framework}
G.~D. Kuruppu Kuruppu~Appuhamilage, M.~Hussain, M.~Zaman, and W.~A. Khan, ``A health digital twin framework for discrete event simulation based optimised critical care workflows,'' \emph{NPJ Digital Medicine}, 2025.

\bibitem{avgousti2020robotic}
S.~Avgousti, P.~Masouras, E.~G. Christoforou \emph{et~al.}, ``Robotic systems in current clinical practice,'' in \emph{2020 21st IEEE Mediterranean Electrotechnical Conference (MELECON)}, 2020.

\bibitem{qian2023deep}
C.~Qian and H.~Ren, ``Deep reinforcement learning in surgical robotics: Enhancing the automation level,'' \emph{arXiv preprint arXiv:2309.00773}, 2023.

\bibitem{bulej2021managing}
L.~Bulej, T.~Bure{\v{s}}, A.~Filandr \emph{et~al.}, ``Managing latency in edge--cloud environment,'' \emph{Journal of Systems \& Software}, 2021.

\bibitem{yang2020bigdata}
C.~Yang, S.~Lan, L.~Wang \emph{et~al.}, ``Big data driven edge-cloud collaboration architecture for cloud manufacturing: A software defined perspective,'' \emph{IEEE Access}, 2020.

\bibitem{sun2022digital}
T.~Sun, X.~He, X.~Song, L.~Shu, and Z.~Li, ``The digital twin in medicine: A key to the future of healthcare?'' \emph{Frontiers in Medicine}, 2022.

\bibitem{seth2024digital}
I.~Seth, B.~Lim, P.~Y.~J. Lu \emph{et~al.}, ``Digital twins use in plastic surgery: A systematic review,'' \emph{Journal of Clinical Medicine}, 2024.

\bibitem{mihai2022digital}
S.~Mihai, M.~Yaqoob, D.~V. Hung \emph{et~al.}, ``Digital twins: A survey on enabling technologies, challenges, trends and future prospects,'' \emph{IEEE Communications Surveys \& Tutorials}, 2022.

\bibitem{ahmed2020potential}
H.~Ahmed and L.~Devoto, ``The potential of a digital twin in surgery,'' \emph{Surgical Innovation}, 2020.

\bibitem{shi2022synergistic}
Y.~Shi, X.~Deng, Y.~Tong \emph{et~al.}, ``Synergistic digital twin and holographic ar-guided percutaneous puncture of respiratory liver tumor,'' \emph{IEEE Transactions on Human-Machine Systems}, 2022.

\bibitem{suraci2025migrate}
C.~Suraci, O.~Chukhno, G.-M. Muntean \emph{et~al.}, ``Migrate or not: Medical digital twins in the era of 6g edge-based networks,'' \emph{IEEE Access}, 2025.

\bibitem{fairley2019improving}
M.~Fairley, D.~Scheinker, and M.~L. Brandeau, ``Improving the efficiency of the operating room environment with an optimization and machine learning model,'' \emph{Health Care Management Science}, 2019.

\bibitem{lu2020digital}
Y.~Lu, C.~Liu, K.~I.-K. Wang, H.~Huang, and X.~Xu, ``Digital twin-driven smart manufacturing: Connotation, reference model, applications and research issues,'' \emph{Robotics and Computer-Integrated Manufacturing}, 2020.

\bibitem{fuller2020digital}
A.~Fuller, Z.~Fan, C.~Day, and C.~Barlow, ``Digital twin: Enabling technologies, challenges and open research,'' \emph{IEEE Access}, 2020.

\bibitem{Glaessgen2012DigitalTwin}
E.~Glaessgen and D.~Stargel, ``The digital twin paradigm for future nasa and u.s. air force vehicles,'' in \emph{53rd AIAA/ASME/ASCE/AHS/ASC Structures, Structural Dynamics and Materials Conference}, 2012.

\bibitem{Grieves2017DTParadigm}
M.~Grieves and J.~Vickers, ``Digital twin: Mitigating unpredictable, undesirable emergent behavior,'' in \emph{Transdisciplinary Perspectives on Complex Systems}.\hskip 1em plus 0.5em minus 0.4em\relax Springer, 2017, pp. 85--113.

\bibitem{Grieves2014Whitepaper}
M.~Grieves, ``Digital twin: Manufacturing excellence through virtual factory replication,'' \emph{White Paper, Florida Institute of Technology}, 2014.

\bibitem{Kritzinger2018Survey}
W.~Kritzinger, M.~Karner, G.~Traar, J.~Henjes, and W.~Sihn, ``Digital twin in manufacturing: A survey on the state of the art,'' in \emph{Procedia CIRP}, vol.~72, 2018, pp. 135--140.

\bibitem{Barricelli2019SurveyDT}
B.~R. Barricelli, E.~Casiraghi, and D.~Fogli, ``A survey on digital twin: Definitions, characteristics, applications, and design implications,'' \emph{IEEE Access}, vol.~7, pp. 167\,653--167\,671, 2019.

\bibitem{Jones2020CharacterisingDT}
D.~Jones, C.~Snider, A.~Nassehi, J.~Yon, and B.~Hicks, ``Characterising the digital twin: A systematic literature review,'' \emph{CIRP Journal of Manufacturing Science and Technology}, vol.~29, pp. 36--52, 2020.

\bibitem{Boschert2016DT}
S.~Boschert and R.~Rosen, ``Digital twin—the simulation aspect,'' in \emph{Mechatronic Futures}.\hskip 1em plus 0.5em minus 0.4em\relax Springer, 2016, pp. 59--74.

\bibitem{Schleich2017ShapingDT}
B.~Schleich, N.~Anwer, L.~Mathieu, and S.~Wartzack, ``Shaping the digital twin for design and production engineering,'' \emph{CIRP Annals}, vol.~66, no.~1, pp. 141--144, 2017.

\bibitem{Tao2018Shopfloor}
F.~Tao and M.~Zhang, ``Digital twin shop-floor: A new shop-floor paradigm towards smart manufacturing,'' \emph{Computers in Industry}, vol. 100, pp. 25--40, 2018.

\bibitem{Zheng2019Framework}
Y.~Zheng, S.~Yang, and H.~Cheng, ``An application framework of digital twin for smart manufacturing,'' \emph{Proceedings of the Institution of Mechanical Engineers, Part B: Journal of Engineering Manufacture}, vol. 233, no.~5, pp. 1342--1351, 2019.

\bibitem{Leng2021DTCPS}
J.~Leng, P.~Jiang, K.~Xu, Q.~Liu, X.~Zhou, and J.~Zhao, ``Digital twin-driven manufacturing cyber–physical systems: Connotation, reference model, applications and research issues,'' \emph{Journal of Manufacturing Systems}, vol.~58, pp. 346--361, 2021.

\bibitem{ISO23247-1-2021}
{International Organization for Standardization}, ``Iso 23247-1:2021 digital twin framework for manufacturing -- part 1: Overview and general principles,'' 2021, standard.

\bibitem{Madni2019MBSE}
A.~M. Madni, C.~A. Madni, and S.~D. Lucero, ``Leveraging digital twin technology in model-based systems engineering,'' \emph{Systems}, vol.~7, no.~1, p.~7, 2019.

\bibitem{Uhlemann2017DTIndustry40}
T.~H.-J. Uhlemann, C.~Lehmann, and R.~Steinhilper, ``The digital twin: Realizing the cyber-physical production system for industry 4.0,'' \emph{Procedia CIRP}, vol.~61, pp. 335--340, 2017.

\bibitem{Negri2017CPSRoles}
E.~Negri, L.~Fumagalli, and M.~Macchi, ``A review of the roles of digital twin in cps-based production systems,'' in \emph{Procedia Manufacturing}, vol.~11, 2017, pp. 939--948.

\bibitem{Minerva2020DTCIoT}
R.~Minerva, A.~Lee, and N.~Crespi, ``Digital twin in the iot context: A survey on technical features, scenarios, and architectural models,'' \emph{Proceedings of the IEEE}, vol. 108, no.~10, pp. 1785--1824, 2020.

\bibitem{ElSaddik2018DTMultimedia}
A.~El~Saddik, ``Digital twins: The convergence of multimedia technologies,'' \emph{IEEE MultiMedia}, vol.~25, no.~2, pp. 87--92, 2018.

\bibitem{Rasheed2020Values}
A.~Rasheed, O.~San, and T.~Kvamsdal, ``Digital twin: Values, challenges and enablers from a modeling perspective,'' \emph{IEEE Access}, vol.~8, pp. 21\,980--22\,012, 2020.

\bibitem{Khan2020DTDefinitions}
W.~Khan, R.~Thakur, G.~Sreedevi, and et~al., ``Digital twin: A comprehensive review of enabling technologies, architecture and challenges,'' \emph{Journal of Industrial Information Integration}, vol.~20, p. 100190, 2020.

\bibitem{Verdouw2021AgriFoodDT}
C.~Verdouw, B.~Tekinerdogan, A.~Beulens, and S.~Wolfert, ``Digital twins in smart farming,'' \emph{Computers and Electronics in Agriculture}, vol. 184, p. 106067, 2021.

\bibitem{Tao2019BookDT}
F.~Tao, Q.~Qi, A.~Liu, and A.~Y.~C. Nee, \emph{Digital Twin Driven Smart Manufacturing}.\hskip 1em plus 0.5em minus 0.4em\relax Academic Press, 2019.

\bibitem{khan202310526}
Z.~Khan, ``10526 the world's first proof of concept of the practical potential in artificially intelligent digital twins in advanced laparoscopic training,'' \emph{Journal of Minimally Invasive Gynecology}, vol.~30, no.~11, p. S127, 2023.

\bibitem{diniz2025digital}
P.~Diniz, B.~Grimm, F.~Garcia \emph{et~al.}, ``Digital twin systems for musculoskeletal applications: A current concepts review,'' \emph{Knee Surgery, Sports Traumatology, Arthroscopy}, 2025.

\bibitem{bonne2022digital}
S.~Bonne, W.~Panitch, K.~Dharmarajan \emph{et~al.}, ``A digital twin framework for telesurgery in the presence of varying network quality of service,'' in \emph{2022 IEEE 18th International Conference on Automation Science and Engineering (CASE)}, 2022.

\bibitem{sapkota2024machine}
M.~S. Sapkota, F.~Doctor, H.~Herrera \emph{et~al.}, ``Machine learning to predict surgery duration: Towards implementing ai and digital twin for effective scheduling,'' in \emph{2024 IEEE International Conference on Medical Artificial Intelligence (MedAI)}, 2024.

\bibitem{khan2022digital}
S.~Khan, T.~Arslan, and T.~Ratnarajah, ``Digital twin perspective of fourth industrial and healthcare revolution,'' \emph{IEEE Access}, 2022.

\bibitem{qamsane2019unified}
Y.~Qamsane, C.-Y. Chen, E.~C. Balta \emph{et~al.}, ``A unified digital twin framework for real-time monitoring and evaluation of smart manufacturing systems,'' in \emph{2019 IEEE 15th International Conference on Automation Science and Engineering (CASE)}, 2019.

\bibitem{yin2019realtime}
S.~Yin, J.~J. Rodriguez-Andina, and Y.~Jiang, ``Real-time monitoring and control of industrial cyberphysical systems: Integrated plant-wide monitoring and control,'' \emph{IEEE Industrial Electronics Magazine}, 2019.

\bibitem{aloqaily2023digital}
M.~Aloqaily, O.~Bouachir, and F.~Karray, ``Digital twin for healthcare immersive services: fundamentals, architectures, and open issues,'' in \emph{Digital Twin for Healthcare}, 2023.

\bibitem{klimo2023digital}
M.~Klimo, M.~Kva{\v{s}}{\v{s}}ay, and N.~Kvassayov{\'a}, ``Digital twin and modelling a 3d human body in healthcare,'' in \emph{2023 21st International Conference on Emerging eLearning Technologies and Applications (ICETA)}, 2023.

\bibitem{coorey2022health}
G.~Coorey, G.~A. Figtree, D.~F. Fletcher \emph{et~al.}, ``The health digital twin to tackle cardiovascular disease---a review of an emerging interdisciplinary field,'' \emph{npj Digital Medicine}, 2022.

\bibitem{badash2016innovations}
I.~Badash, K.~Burtt, C.~A. Solorzano, and J.~N. Carey, ``Innovations in surgery simulation: a review of past, current and future techniques,'' \emph{Annals of Translational Medicine}, 2016.

\bibitem{huang2019telemedicine}
E.~Y. Huang, S.~Knight, C.~R. Guetter \emph{et~al.}, ``Telemedicine and telementoring in the surgical specialties: A narrative review,'' \emph{The American Journal of Surgery}, 2019.

\bibitem{gross1998computer}
M.~H. Gross, ``Computer graphics in medicine: From visualization to surgery simulation,'' \emph{Computer Graphics}, 1998.

\bibitem{chen2007human}
J.~Y.~C. Chen, E.~C. Haas, and M.~J. Barnes, ``Human performance issues and user interface design for teleoperated robots,'' \emph{IEEE Transactions on Systems, Man, and Cybernetics, Part C (Applications and Reviews)}, 2007.

\bibitem{rajkomar2019machine}
A.~Rajkomar, J.~Dean, and I.~Kohane, ``Machine learning in medicine,'' \emph{The New England Journal of Medicine}, 2019.

\bibitem{kamel2021digital}
M.~N. Kamel~Boulos and P.~Zhang, ``Digital twins: From personalised medicine to precision public health,'' \emph{Journal of Personalized Medicine}, 2021.

\bibitem{qi2018digital}
Q.~Qi and F.~Tao, ``Digital twin and big data towards smart manufacturing and industry 4.0: 360 degree comparison,'' \emph{IEEE Access}, 2018.

\bibitem{wang2021digital}
P.~Wang and M.~Luo, ``A digital twin-based big data virtual and real fusion learning reference framework supported by industrial internet towards smart manufacturing,'' \emph{Journal of Manufacturing Systems}, 2021.

\bibitem{alauddin2018bibliometric}
M.~Alauddin, F.~Khan, S.~Imtiaz, and S.~Ahmed, ``A bibliometric review and analysis of data-driven fault detection and diagnosis methods for process systems,'' \emph{Industrial \& Engineering Chemistry Research}, 2018.

\bibitem{rong2020ai}
G.~Rong, A.~Mendez, E.~Bou~Assi, B.~Zhao, and M.~Sawan, ``Artificial intelligence in healthcare: Review and prediction case studies,'' \emph{Engineering}, 2020.

\bibitem{tseng2020future}
M.-L. Tseng, C.-H. Chang, C.-W.~R. Lin \emph{et~al.}, ``Future trends and guidance for the triple bottom line and sustainability: a data driven bibliometric analysis,'' \emph{Environmental Science and Pollution Research}, 2020.

\bibitem{bagheriye2025advancements}
L.~Bagheriye and J.~Kwisthout, ``Advancements in real-time oncology diagnosis: Harnessing ai and image fusion techniques,'' \emph{arXiv preprint arXiv:2503.11332}, 2025.

\bibitem{santoso2024maximizing}
A.~Santoso and Y.~Surya, ``Maximizing decision efficiency with edge-based ai systems,'' \emph{Quarterly Journal of Emerging Technologies}, 2024.

\bibitem{rudnicka2024cardiac}
Z.~Rudnicka, K.~Proniewska, M.~Perkins, and A.~Pregowska, ``Cardiac healthcare digital twins supported by ai-based algorithms and extended reality—a systematic review,'' \emph{Electronics}, 2024.

\bibitem{cockrell2024weabm}
C.~Cockrell, Y.~Vodovotz, R.~Zamora, and G.~An, ``The wound environment agent-based model (weabm): a digital twin platform for volumetric muscle loss,'' \emph{bioRxiv}, 2024.

\bibitem{reza2024assessing}
S.~Reza, B.~J. Kovarovic, D.~Bluestein, and P.~D. Bluestein, ``Assessing post-tavr cardiac conduction abnormalities risk using a digital twin of a beating heart,'' \emph{medRxiv}, 2024.

\bibitem{tortora2025medical}
M.~Tortora, F.~Pacchiano, S.~F. Ferraciolli \emph{et~al.}, ``Medical digital twin: A review on technical principles and clinical applications,'' \emph{Journal of Clinical Medicine}, 2025.

\bibitem{halpern2025advances}
G.~A. Halpern, M.~Nemet, D.~M. Gowda \emph{et~al.}, ``Advances and utility of digital twins in critical care and acute care medicine: a narrative review,'' \emph{J Yeungnam Med Sci}, 2025.

\bibitem{kabir2025digital}
M.~R. Kabir, F.~S. Shishir, S.~Shomaji, and S.~Ray, ``Digital twins in healthcare iot: A systematic review,'' \emph{High-Confidence Computing}, 2025.

\bibitem{ringeval2025advancing}
M.~Ringeval, F.~A. Etindele~Sosso, M.~Cousineau, and G.~Par{\'e}, ``Advancing health care with digital twins: Meta-review of applications and implementation challenges,'' \emph{Journal of Medical Internet Research}, 2025.

\bibitem{sel2025survey}
K.~Sel, A.~Hawkins-Daarud, A.~Chaudhuri \emph{et~al.}, ``Survey and perspective on verification, validation, and uncertainty quantification of digital twins for precision medicine,'' \emph{NPJ Digital Medicine}, 2025.

\bibitem{sendak2020path}
M.~P. Sendak, J.~D'Arcy, S.~Kashyap \emph{et~al.}, ``A path for translation of machine learning products into healthcare delivery,'' \emph{EMJ Innov}, 2020.

\bibitem{morley2020ethics}
J.~Morley, C.~C.~V. Machado, C.~Burr \emph{et~al.}, ``The ethics of ai in health care: A mapping review,'' \emph{Social Science \& Medicine}, 2020.

\bibitem{leslie2019understanding}
D.~Leslie, ``Understanding artificial intelligence ethics and safety: A guide for the responsible design and implementation of ai systems in the public sector,'' \emph{The Alan Turing Institute}, 2019.

\bibitem{bhatia2016temporal}
M.~Bhatia and S.~K. Sood, ``Temporal informative analysis in smart-icu monitoring: M-healthcare perspective,'' \emph{Journal of Medical Systems}, 2016.

\bibitem{aydin2022openai}
{\"O}.~Ayd{\i}n and E.~Karaarslan, ``Openai chatgpt generated literature review: Digital twin in healthcare,'' in \emph{Emerging Computer Technologies 2}, 2022.

\bibitem{cellina2023digital}
M.~Cellina, M.~C{\`e}, M.~Al{\`i} \emph{et~al.}, ``Digital twins: The new frontier for personalized medicine?'' \emph{Applied Sciences}, 2023.

\end{thebibliography}


\end{document}